\documentclass[11pt]{article} 
\usepackage{graphicx}
\usepackage[width=145mm,top=45mm,bottom=45mm]{geometry}
\usepackage{amssymb,amsmath} % e.g. mathbb
\usepackage{bbm}

\usepackage{mathtools} % e.g. equations; loads amsmath
\usepackage{amsthm} % proof environment
\usepackage{IEEEtrantools}
\usepackage{dsfont}
\usepackage{tikz}
\usepackage{float}
\usepackage{todonotes}
\usepackage{authblk}
\usetikzlibrary{arrows.meta, positioning}
\usepackage[most]{tcolorbox}

\definecolor{hanblue}{rgb}{0.27, 0.42, 0.81}
\definecolor{mordantred19}{rgb}{0.68, 0.05, 0.0}
\definecolor{red}{rgb}{0.68, 0.05, 0.0}
\definecolor{green}{rgb}{0.0, 0.5, 0.0}

\definecolor{prooflinecolor}{gray}{0.6}

\usepackage[colorlinks, citecolor=hanblue,linkcolor=green]{hyperref}
\usepackage[capitalise]{cleveref}

\tcbuselibrary{breakable}
\usepackage{stmaryrd} % norm |||.|||
\usepackage{todonotes}
\usepackage{comment}
\usepackage{color}
\usepackage{subcaption}
\usepackage{mathabx}

\theoremstyle{plain}

\newtheorem{theorem}{Theorem}[section]
\newtheorem{lemma}[theorem]{Lemma}
\newtheorem{corollary}[theorem]{Corollary}

\newtheorem{definition}[theorem]{Definition}

\newtheorem{remark}[theorem]{Remark}

\renewcommand{\u}{\mathbf{u}}
\newcommand{\f}{\mathbf{f}}
\newcommand{\x}{\mathbf{x}}
\newcommand{\y}{\mathbf{y}}
\newcommand{\w}{\mathbf{w}}

\newcommand{\U}{\mathbf{U}}
\newcommand{\F}{\mathbf{F}}
\newcommand{\X}{\mathbf{X}}
\newcommand{\Y}{\mathbf{Y}}
\newcommand{\W}{\mathbf{W}}
\newcommand{\R}{\mathbb{R}}

\newcommand{\uF}{\widehat{F}}
\newcommand{\lF}{\widecheck{F}}
\newcommand{\uG}{\widehat{G}}
\newcommand{\lG}{\widecheck{G}}

\usepackage{tikz}
\usetikzlibrary{positioning, arrows.meta, fit, calc, backgrounds}
\usepackage{caption}
\usepackage{xcolor}
\usetikzlibrary{positioning, arrows.meta}
\definecolor{boxgreen}{RGB}{204, 255, 204}      % Light green for "step of flow"
\definecolor{bggreen}{RGB}{210, 255, 210}       % Slightly different green for background (a)
\definecolor{boxyellow}{RGB}{255, 248, 220}     % Light yellow for "squeeze"
\definecolor{bordergray}{RGB}{130, 130, 130}    % Gray for borders
\usepackage[most]{tcolorbox}
\tcbuselibrary{breakable}

\title{An invertible generative model for forward and inverse problems}
\author[2]{Christoph Brune}
\author[2]{Marcello Carioni}
\author[1,3]{Tristan van Leeuwen}
\author[1]{Lasse Veenstra}
\affil[1]{Centrum Wiskunde en Informatica, Amsterdam, The Netherlands}
\affil[2]{University of Twente, Enschede, The Netherlands}
\affil[3]{Utrecht University, Utrecht, The Netherlands}

\date{}
\begin{document}

\maketitle

% REQUIRED
\begin{abstract}
We formulate inverse problems in a Bayesian framework and aim to train an invertible generative model that is capable of simulation (i.e., sampling from the likelihood) and inference (i.e., sampling from the posterior).  We call such a generative model a \emph{Reversible Simulator}. We first construct an explicit instance of a reversible simulator by combining lower and upper triangular normalizing flows associated with conditional sampling into a single invertible map. We then establish basic structural properties of this construction, investigate its non-uniqueness, and derive a variational formulation that enables the generative model to be trained directly from paired samples.
Finally, we illustrate the proposed framework on analytically tractable and stylized numerical examples, demonstrating its potential as a unified approach to conditional generative modeling for forward and inverse problems.
\end{abstract}

% REQUIRED
\textbf{Keywords.} Inverse problems, Bayesian Inference, Generative Models, Optimal Transport

% REQUIRED
\textbf{MSCcodes.}
62F15, 49Q22,  68T07

%%%%%%%%%%%%%%%%%%%%%%%%%%%%%%%%%%%%%%%%%%%%%%%%%%%%%%%%%%%%%%%%%%%%%%%%%%%%%%%%%%%%%%%%%%%%%%%%%%%%%%%%%%%
%%%%%%%%%%%%%%%%%%%%%%%%%%%%%%%%%%%%%%%%%%%%%%%%%%%%%%%%%%%%%%%%%%%%%%%%%%%%%%%%%%%%%%%%%%%%%%%%%%%%%%%%%%%
\section{Introduction}
%%%%%%%%%%%%%%%%%%%%%%%%%%%%%%%%%%%%%%%%%%%%%%%%%%%%%%%%%%%%%%%%%%%%%%%%%%%%%%%%%%%%%%%%%%%%%%%%%%%%%%%%%%%
%%%%%%%%%%%%%%%%%%%%%%%%%%%%%%%%%%%%%%%%%%%%%%%%%%%%%%%%%%%%%%%%%%%%%%%%%%%%%%%%%%%%%%%%%%%%%%%%%%%%%%%%%%%
Inverse problems occur in many applications and are typically formulated in terms of a \emph{forward operator}, $K : \mathbb{R}^{n}\to \mathbb{R}^m$, \emph{measurements}, $\f\in\mathbb{R}^m$, and entails finding $\u\in\mathbb{R}^n$ such that $K\u=\f$. 
This problem is typically ill-posed as an inverse of $K$ does not exist or at best is highly unstable. In some cases, when the problem is only mildly ill-posed, it is possible to derive an explicit expression for a regularized inverse $R\approx K^\dagger$ which can also be efficiently evaluated. A prime example is the filtered back-projection method for CT reconstruction \cite[and references therein]{hansen2021computed} or the Wiener filter for image deblurring \cite[and references therein]{gunturk2018image}. While computationally very efficient, such explicit constructions typically require fully sampled measurements. Thus, measurements take a long time, but computation is relatively cheap. 

There is therefore considerable interest in shortening the acquisition process, often by collecting fewer or undersampled measurements. In such settings, direct inversion methods may no longer be applicable or may produce unsatisfactory reconstructions. A common alternative is to formulate the inverse problem as a \emph{variational problem}:
\begin{equation}\label{eq:variational}
\min_{\u\in\mathbb{R}^n} \mathcal{D}(K\u,\f)+\mathcal{R}(\u), 
\end{equation}
with $\mathcal{D}:\mathbb{R}^{m}\times\mathbb{R}^m\to\mathbb{R}_+$ the \emph{data-fidelity} and $\mathcal{R}:\mathbb{R}^n\to\mathbb{R}^+$ the \emph{regularization term}. This implicitly defines an inverse mapping $R:\mathbb{R}^m\to\mathbb{R}^n$ and provides a powerful framework for analysis and algorithm design \cite{scherzer2009variational}. The resulting optimization problem is often cumbersome to solve, requiring many evaluations of $K$ to converge using an iterative scheme. Consequently, while undersampling reduces acquisition time, it often shifts the computational burden to the reconstruction stage.

\emph{Data-driven approaches} aim at closing the gap in computational efficiency between direct and variational methods by learning a mapping $R:\mathbb{R}^m\to\mathbb{R}^n$ from example data $\{(\u_i,\f_i)\}_{i=1}^N$ \cite{arridge2019solving}. When the learned inverse can be evaluated efficiently, one can use it for fast inference at the expense of a computationally intensive off-line training phase. Care needs to be taken here with the inherent ill-posedness; if the training data contains samples for which $\|\f_i - \f_j\| / \|\u_i - \u_j\|$ is small, the resulting map $R$ may be highly ill-conditioned.
Despite that, notable examples have achieved impressive reconstruction results for a wide range of inverse problems \cite{adler2018learned, jin2017deep, gilton2021deep, monga2021algorithm, zhu2018image, mukherjee2021end, li2020nett, habring2024neural}. 

A different perspective is given by the \emph{Bayesian framework} in which $\u,\f$ are interpreted as realizations of random variables, $\U,\F$, that follow some underlying distribution \cite{tarantola}. Typically, one formulates a \emph{prior} distribution (defining a probability measure for $\U$) and a likelihood (defining a conditional probability for $\F$ given $\U=\u$). As per Bayes' rule, this results in a \emph{posterior} distribution, giving a conditional probability measure of $\U$ given $\F=\f$. Surprisingly, this Bayesian formulation is typically well-posed \cite{Latz}. In a way, the posterior is the ultimate answer to the inverse problem, as it captures assumptions on the data-generating process and prior information and allows one to give a probabilistic answer. Characterising the posterior in general is difficult, and one often results to either estimating the mode (the MAP estimate) and local variance or 
drawing samples from it using MCMC techniques. MAP-estimation can be challenging for high-dimensional problems, although matrix-sketching techniques have been developed to push the boundaries of this approach \cite{ghattas2024}. MCMC-sampling techniques, on the other hand, are a powerful way to more accurately characterize complicated probability distributions, but suffer from the curse of dimensionality. Alleviating this is a very active field of research with many success stories \cite{NEURIPS2018_335cd1b9}. 

\emph{Generative models} \cite{goodfellow2020generative, kingma2013auto, kingma2016improved, patrini2020sinkhorn, NEURIPS2018_69386f6b} adapt well to Bayesian frameworks, since they can generate samples from complicated probability distributions efficiently. In particular, they are well suited  to address Bayesian formulations of inverse problems \cite{adler2018deep,  oliviero2025generative, goh2019solving}. 
Moreover, they have also been applied successfully to uncertainty quantification in inverse problems \cite{adler2018deep,Radev2022,Radev2023, repetti2019scalable}. The basic gist of these methods is that they are trained in an off-line phase based on training data and are then deployed to generate samples from the corresponding posterior efficiently. Generally, this leads to a map $R:\mathbb{R}^{n+m}\to\mathbb{R}^{m}$ that will map samples from a reference distribution to samples from the posterior, i.e. $\U = R(\X,\f)$ with $\X$ distributed according to a reference distribution over $\mathbb{R}^n$ (i.e., standard normal) will result in $\U$ being distributed according to the posterior conditioned on $\f$ \cite{Ardizzone2018}. Similarly, one may train a generative model to sample from the likelihood distribution, yielding a map $S:\mathbb{R}^{n+m}\to\mathbb{R}^{n}$ such that $\F=S(\Y,\u)$ with $\Y$ distributed according to a reference distribution over $\mathbb{R}^m$ will result in $\F$ being distributed according to the likelihood conditioned on $\u$.

\begin{figure}
\centering
\includegraphics[scale=.3]{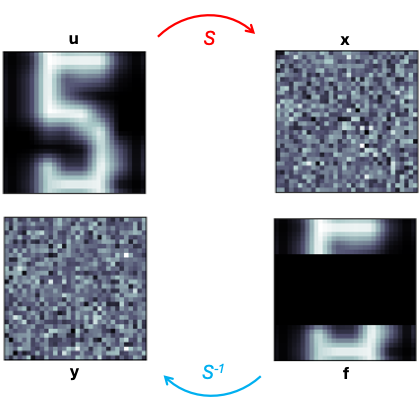}
\caption{Schematic depiction of the map $S$ for an inpainting problem: the simulation entails removing part of the digit; the inverse of the map produces an inpainted digit.}\label{fig:intro}
\end{figure}
%%%%%%%%%%%%%%%%%%%%%%%%%%%%%%%%%%%%%%%%%%%%%%%%%%%%%%%%%%%%%%%%%%%%%%%%%%%%%%%%%%%%%%%%%%%%%%%%%%%%%%%%%%%

\subsection{Contributions of the paper}
%%%%%%%%%%%%%%%%%%%%%%%%%%%%%%%%%%%%%%%%%%%%%%%%%%%%%%%%%%%%%%%%%%%%%%%%%%%%%%%%%%%%%%%%%%%%%%%%%%%%%%%%%%%
This work proposes a novel framework for the use of generative models in inverse problems, which unifies the tasks of simulation (sampling from the likelihood) and inference (sampling from the posterior). 
In standard approaches, these two tasks are typically handled by separate conditional generative models, even though they describe complementary aspects of the same joint distribution. Representing them through a single invertible map therefore provides a natural mechanism for enforcing consistency between the learned forward and inverse models. Moreover, the shared representation allows paired training data to constrain both tasks simultaneously, potentially reducing the computational costs associated with training two separate models. Such a unified model is particularly appealing in applications in which simulation and inference must be repeatedly alternated, including uncertainty quantification, model validation, simulation-based inference and experimental design.

The central object in this framework is a single invertible map
\begin{align}
S:\mathbb{R}^{n+m}\to\mathbb{R}^{n+m},
\end{align}
which can be evaluated in the forward direction to simulate observations and in the reverse direction to generate posterior samples; we will call such map $S$ a \emph{Reversible Simulator}. More precisely, the map $S$ can be learned from training data $\{(\u_i,\f_i)\}_{i=1}^N$ and defines a generative model for simulation and inference in the sense that for given $\u\in\mathbb{R}^n$, $\f\in\mathbb{R}^m$ and $\X,\Y$ distributed according to some reference distribution the application of $S$
\begin{equation}
(\X',\F) = S(\u,\Y), 
\end{equation}
yields $\F$ distributed according to the likelihood, conditioned on $\u$ and the application of $S^{-1}$
\begin{equation}
(\U,\Y')=S^{-1}(\X,\f), 
\end{equation}
yields $\U$ distributed according to the posterior conditioned on $\f$. Furthermore, $\X', \Y'$ are distributed according to their respective reference distributions. A schematic depiction is shown in Figure \ref{fig:intro}. The overarching goal of this paper is to establish a theoretical framework
for reversible simulators by investigating their existence, uniqueness and
explicit construction, and developing a principled basis for learning them from
data.

In particular, the main contributions of the paper are as follows: 

\begin{itemize}
    \item We provide an explicit construction of a reversible simulator by
    combining conditional normalizing flows represented by lower- and
    upper-triangular maps.
    \item We perform a theoretical analysis regarding the uniqueness of reversible simulators.
    \item We derive a variational formulation based on training objectives
    which can be evaluated directly from paired samples
    $\{(\u_i,\f_i)\}_{i=1}^N$.
    \item We present analytical and numerical examples illustrating the
    construction and the potential advantages of the proposed generative
    framework.
 \end{itemize}

%%%%%%%%%%%%%%%%%%%%%%%%%%%%%%%%%%%%%%%%%%%%%%%%%%%%%%%%%%%%%%%%%%%%%%%%%%%%%%%%%%%%%%%%%%%%%%%%%%%%%%%%%%%

\subsection{Related work}
%%%%%%%%%%%%%%%%%%%%%%%%%%%%%%%%%%%%%%%%%%%%%%%%%%%%%%%%%%%%%%%%%%%%%%%%%%%%%%%%%%%%%%%%%%%%%%%%%%%%%%%%%%%
Closest in spirit to this work is the work by \cite{Ardizzone2018} which first proposed to use invertible neural networks for solving inverse problems. Translated to the notation of this paper, they treat the case $m \leq n$ and use an invertible neural network to learn a mapping $S : \mathbb{R}^{n}\rightarrow \mathbb{R}^{m+k}$ with $k = n-m$. For given data $\f$ and a random sample from latent space $\x$, this mapping is supposed to produce a sample from the posterior $\u = S(\f,\x)$. This will work in case the underlying likelihood collapses on to the manifold $\f = K\u$ (i.e., noiseless measurements) and will thus sample from the prior restricted to this manifold. This approach using generic invertible architectures was later superseded by the use of normalizing flows for amortized Bayesian inference \cite{Marzouk2017,Radev2022,Radev2023,orozco2025aspire, oliviero2025generative}, as it provides a more principled way of doing so.
This work can be seen as a bridge between these two approaches and provides a unified framework for using invertible mappings for simulation and inference.

Worth mentioning also is the work by \cite{moens2022viscos} which proposes a generic framework for conditioning a given normalizing flow on any of its input variables. In principle, this approach could be applied to use a generic normalizing flow that represents the distribution of $(\U,\F)$ for either simulation (by conditioning on $\U=\u$) or inference (by conditioning on $\F=\f$). Computationally this is expensive though as for each sample a non-linear system of equations needs to be solved. Moreover, the construction is not generic in the sense that the normalizing flow needs to capture the required dependencies. The current work explicitly constructs a mapping that captures these dependencies and allows for sampling of the two conditional distributions more directly. This comes at the cost of having to fix beforehand on which variables one wishes to be able to condition.

We also mention that conditional generative models used to learn undetermined inverse problems reconstructions are widely available in the literature as variants of popular generative models. Examples are conditional normalizing flows \cite{winkler2019learning, batzolis2021caflow,  lugmayr2020srflow}, conditional diffusion models \cite{batzolis2021conditional, rasul2021autoregressive}, conditional GANs \cite{mirza2014conditional}.
Moreover, generative models have been recently used in form of Plug-and-Play models for inverse problems reconstructions \cite{kamilov2017plug, romano2017little, carioni2024unsupervised, martin2025pnp}.
Finally, it is worth mentioning the latest developments in non-deterministic generative models such as stochastic normalising flows \cite{wu2020stochastic, hagemann2022stochastic}, stochastic interpolants \cite{albergo2023stochastic}, and diffusion models \cite{song2020score, batzolis2021conditional}. We hypothesize that the proposed framework of mapping between $(\U,\Y)$ and $(\X,\F)$ can be implemented using any of these generative models (see also section \ref{conclusion}).
%%%%%%%%%%%%%%%%%%%%%%%%%%%%%%%%%%%%%%%%%%%%%%%%%%%%%%%%%%%%%%%%%%%%%%%%%%%%%%%%%%%%%%%%%%%%%%%%%%%%%%%%%%%

\subsection{Outline}

The remainder of the paper is organized as follows. We first review some necessary notation and preliminaries on triangular transport maps in \cref{prelim}. The main results regarding the Reversible Simulator are presented in \cref{main} and numerical examples are given in \cref{numerics}. Finally, \cref{conclusion} concludes the paper and discuss further developments.

%%%%%%%%%%%%%%%%%%%%%%%%%%%%%%%%%%%%%%%%%%%%%%%%%%%%%%%%%%%%%%%%%%%%%%%%%%%%%%%%%%%%%%%%%%%%%%%%%%%%%%%%%%%
%%%%%%%%%%%%%%%%%%%%%%%%%%%%%%%%%%%%%%%%%%%%%%%%%%%%%%%%%%%%%%%%%%%%%%%%%%%%%%%%%%%%%%%%%%%%%%%%%%%%%%%%%%%
\section{Preliminaries}\label{prelim}
%%%%%%%%%%%%%%%%%%%%%%%%%%%%%%%%%%%%%%%%%%%%%%%%%%%%%%%%%%%%%%%%%%%%%%%%%%%%%%%%%%%%%%%%%%%%%%%%%%%%%%%%%%%
%%%%%%%%%%%%%%%%%%%%%%%%%%%%%%%%%%%%%%%%%%%%%%%%%%%%%%%%%%%%%%%%%%%%%%%%%%%%%%%%%%%%%%%%%%%%%%%%%%%%%%%%%%%

%%%%%%%%%%%%%%%%%%%%%%%%%%%%%%%%%%%%%%%%%%%%%%%%%%%%%%%%%%%%%%%%%%%%%%%%%%%%%%%%%%%%%%%%%%%%%%%%%%%%%%%%%%%
\subsection{Notation}
%%%%%%%%%%%%%%%%%%%%%%%%%%%%%%%%%%%%%%%%%%%%%%%%%%%%%%%%%%%%%%%%%%%%%%%%%%%%%%%%%%%%%%%%%%%%%%%%%%%%%%%%%%%
Vectors will be denoted by boldface symbols, e.g., $\u\in\mathbb{R}^n, \f\in\mathbb{R}^m$ and operators by italic capitals, e.g., $K:\mathbb{R}^n\to\mathbb{R}^m$ and $S:\mathbb{R}^{n+m}\to\mathbb{R}^{n+m}$.
Random variables will be denoted by bold capitals, e.g. $\U,\F$. We denote the probability measure underlying a random variable $\U$ by $\mu_\U$. The joint distribution of two variables $(\U,\F)$ will be denoted similarly as $\mu_{\U,\F}$. When the distribution is absolutely continuous, we denote the corresponding density as $\pi_\U$ etc. For sake of generality, however, and to handle singular data distributions (e.g., data concentrated on manifolds) we will formulate the results preferably in terms of the measures. We denote by $\mathbb{E}_{\w \sim \mu_\W}$ the expectation with respect to a probability measure $\mu_\W$. With the goal of ease notation we will often write $\mathbb{E}_{\mu_\W}$ instead of $\mathbb{E}_{\w \sim \mu_\W}$. The variable dependence will be clear from the context. Furthermore, we denote the push-forward of a measure by the map $T$ by $T_\#\mu = \mu\circ T^{-1}$. When applicable, this is denoted in terms of the corresponding density as $T_\#\pi = (\pi \circ T^{-1})|\nabla T^{-1}|$. Similarly, the pull-back is denoted as $T^\#\mu = \mu\circ T$ and $T^\#\pi = (\pi\circ T) |\nabla T|$.

The main objects of interest in this paper are the joint distribution $\mu_{\U,\F}$ over $\mathbb{R}^{n+m}$ which describes the relation between $\U$ (taking values in $\mathbb{R}^n$) and $\F$ (taking values in $\mathbb{R}^m$), its conditionals $\mu_{\F|\U}$ (also referred to as the likelihood) and $\mu_{\U|\F}$ (also referred to as the posterior), and its marginals $\mu_\U$ (referred to as the prior in the inverse problems literature) and $\mu_\F$.
Thanks to the existence of conditional expectations (equivalently by existence of a disintegration) we can write
\begin{align*}
\mu_{\U,\F} = \mu_{\F|\U} \otimes \mu_{\U} = \mu_{\U|\F} \otimes \mu_{\F},
\end{align*}
where the first equality is intended rigorously as 
\begin{align*}
    \mathbb{E}_{(\u,\f) \sim \mu_{\U,\F}} \varphi(\u,\f) = \mathbb{E}_{\u \sim \mu_\U}\mathbb{E}_{\f \sim \mu_{\F|\U=\u}} \varphi(\u,\f) 
\end{align*}
for every test function $\varphi:\mathbb{R}^{n\times m}\to\mathbb{R}$ and the second equality is similarly defined.

%%%%%%%%%%%%%%%%%%%%%%%%%%%%%%%%%%%%%%%%%%%%%%%%%%%%%%%%%%%%%%%%%%%%%%%%%%%%%%%%%%%%%%%%%%%%%%%%%%%%%%%%%%%
\subsection{Conditional sampling using triangular maps}
%%%%%%%%%%%%%%%%%%%%%%%%%%%%%%%%%%%%%%%%%%%%%%%%%%%%%%%%%%%%%%%%%%%%%%%%%%%%%%%%%%%%%%%%%%%%%%%%%%%%%%%%%%%
Our starting point is the construction of an invertible mapping $F$ which represents the joint distribution as
\begin{equation}\label{eq:pushforwardF}
\mu_{\U,\F} = F_\# \mu_{\X,\Y},
\end{equation}
with $\mu_{\X,\Y}$ a given reference distribution. Throughout we assume that the reference distribution is separable, i.e., $\mu_{\X,\Y} \equiv \mu_\X \otimes \mu_\Y$ where $\mu_\X$ and $\mu_\Y$ are the reference distributions over the latent space. Here $\X$ is used to denote the latent variable corresponding to $\U$, and $\Y$ the latent variable corresponding to $\F$.

While there are infinitely many maps that will achieve this, triangular maps have the appealing property that they allow for relatively straightforward evaluation of the conditional distributions $\mu_{\F|\U}$ resp. $\mu_{\U|\F}$ when $F$ exhibits lower resp. upper triangular structure. This triangular structure is also known as the Knothe-Rosenblatt (KR) rearrangement, and correspondingly a triangular transport map is often referred to as the KR-map. It exists under mild conditions on the distributions. For the sake of completeness we reproduce here the main results required (see e.g., \cite[Lemma 1]{Marzouk2017}).
\begin{definition}[Invertible lower-triangular maps]\label{def:invlt}
A map $F :\mathbb{R}^{n+m}\to\mathbb{R}^{n+m}$ is called lower-triangular if its Jacobian $\nabla F \in \mathbb{R}^{(n+m)\times (n+m)}$ is a lower-triangular matrix. Splitting the variables, it can be expressed as 
$$F(\x,\y) = (F_1(\x), F_2(\x,\y)),$$
with $F_1:\mathbb{R}^{n}\to\mathbb{R}^n$ and $F_2:\mathbb{R}^{n+m}\to\mathbb{R}^m$.
We say that $F$ is an invertible lower-triangular map if $F$ is invertible and  its inverse $G = F^{-1}$ is lower-triangular, i.e. given by
$$G(\u,\f) = (G_1(\u), G_2(\u,\f)).$$
Note that this implies that $G_1=F_1^{-1}$, and $G_2$ satisfying
$$G_2(\u,F_2(G_1(\u),\y))=\y,$$ 
and 
\begin{align}\label{eq:invtri}
F_2(\x,G_2(F_1(\x),\f))=\f.
\end{align}
\end{definition}

\begin{definition}[Invertible upper-triangular maps]\label{def:invut}
A map $F :\mathbb{R}^{n+m}\to\mathbb{R}^{n+m}$ is called upper-triangular if its Jacobian $\nabla F \in \mathbb{R}^{(n+m)\times (n+m)}$ is an upper-triangular matrix. Splitting the variables, it can be expressed as 
$$F(\x,\y) = (F_1(\x,\y), F_2(\y)),$$
with $F_1:\mathbb{R}^{n+m}\to\mathbb{R}^n$ and $F_2:\mathbb{R}^{m}\to\mathbb{R}^m$.
We say that $F$ is an invertible upper-triangular map if $F$ is invertible and  its inverse $G = F^{-1}$ is upper-triangular, i.e. given by
$$G(\u,\f) = (G_1(\u,\f), G_2(\f)).$$
Note that this implies that $G_2=F_2^{-1}$, and $G_1$ satisfying
$$G_1(F_1(\x,G_2(\f)),\f)=\x,$$ 
and 
$$F_1(G_1(\u, F_2(\y)), \y)=\u.$$
\end{definition}

The triangular structure of the map $F$ can be exploited to condition on either $\u$ or $\f$, as per the following two lemmas.

\begin{lemma}\label{lemma:like}
Let $\lF : \mathbb{R}^{n+m} \to \mathbb{R}^{n+m}$ be an invertible lower-triangular map with inverse $\lG$
satisfying
$$
\lF_\# \left(\mu_\X \otimes \mu_\Y\right) = \mu_{\U,\F},
$$
for given reference and target measures $\mu_\X, \mu_\Y$, $\mu_{\U,\F}$.
Then, for any $\u \in \mathbb{R}^n$, the map
$$
F_{\text{like}}(\cdot; \u) := \lF_2(\lF_1^{-1}(\u), \cdot),
$$
with inverse
$$F_\text{like}^{-1}(\cdot;\u) := \lG_2(\u,\cdot),$$
pushes forward $\mu_\Y$ to the conditional measure $\mu_{\F|\U=\u}$, i.e.,
$$
\mu_{\F|\U=\u} = F_{\text{like}}(\cdot; \u)_\# \mu_\Y.
$$
\end{lemma}

\begin{figure}
\centering
\begin{tabular}{cc}
\includegraphics[scale=.25]{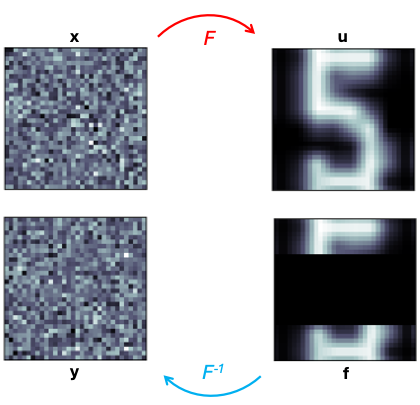}&
\includegraphics[scale=.25]{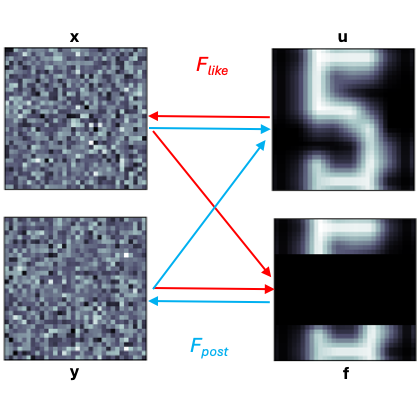}
\end{tabular}
\caption{Schematic depiction of the maps $F$, $F_\text{like}$, and $F_\text{post}$ for an inpainting problem (cf. figure \ref{fig:intro}). A generic map $F$ will map between $(\x,\y)$ and $(\u,\f)$ (left panel) while the triangular structure allows us to perform conditional sampling $(\u,\y)\mapsto \f$ or $(\x,\f)\mapsto \u$ by either forward or back substitution (right panel).}\label{fig:intro2}
\end{figure}

\begin{proof}
We have that $\mu_{\U,\F} = \lF_\# \left(\mu_\X \otimes \mu_\Y\right)$, where $\lF$ is invertible and lower-triangular, that is 
$$
\lF(\x, \y) = (\lF_1(\x), \lF_2(\x, \y)).
$$
We first prove that $(\lF_1)_{\#} \mu_{\X} = \mu_\U$. Indeed, by choosing the test function $\varphi(\u,\f) = \psi(\u)$ we obtain that
    \begin{align*}   \mathbb{E}_{\mu_{\U}}\psi(\u) = \mathbb{E}_{\mu_{(\U,\F)}}\varphi(\u,\f) = \mathbb{E}_{\mu_{(\X,\Y)}}\psi(\lF_1(\x)) = \mathbb{E}_{\mu_{\X}}\psi(\lF_1(\x))
    \end{align*}
    where we used that $\mu_\U$ and $\mu_\X$ are the marginals of $\pi_{\U,\F}$ and $\pi_{\X,\Y}$ respectively. Moreover, since $\lF_1$ is invertible, it also holds that $(\lF_1^{-1})_\# \mu_\U = \mu_\X.$
    \\
By existence of the conditional expectation, we can write $\mu_{\U,\F}$ as
$\mu_{\U,\F} = \mu_{\F|\U} \otimes \mu_\U$.
To prove that $\mu_{\F|\U=\u} = F_\text{like}(\cdot;\u)_\# \mu_\Y$, we will show that the action of $\mu_{\F|\U=\u}$ on test functions matches that of pushing forward $\mu_\Y$ through $F_\text{like}$. Let $\varphi : \mathbb{R}^m \to \mathbb{R}$ be any test function. Then
\begin{align*}
&\mathbb{E}_{\mu_U}\mathbb{E}_{F_\text{like}(\cdot;\u)_\# \mu_\Y} \varphi(\u,\f) = \mathbb{E}_{\mu_U}\mathbb{E}_{\mu_\Y} \varphi(\u,F_\text{like}(\u,\y))
= \mathbb{E}_{\mu_U}\mathbb{E}_{\mu_\Y} \varphi(\u,\lF_2(\lF_1^{-1}(\u), \y))=\\
&\mathbb{E}_{\mu_\U}\mathbb{E}_{\mu_\Y} \varphi(\lF_1(\lF_1^{-1}(\u)),\lF_2(\lF_1^{-1}(\u), \y))
 =  \mathbb{E}_{\mu_\X}\mathbb{E}_{\mu_\Y} \varphi(\lF_1(\x),\lF_2(\x, \y))
 = \mathbb{E}_{\mu_{(\U,\F)}}\varphi(\u,\f),
\end{align*}
where we have used that $(\lF_1^{-1})_\# \mu_\U = \mu_\X$, $\mu_{\U,\F} = \lF_\# \mu_{\X,\Y}$ and the definition of $\lF$. From the previous chain of inequalities and the uniqueness of the conditional expectation it follows that $\mu_{\F|\U=\u} = F_\text{like}(\cdot;\u)_\# \mu_\Y$ as we wanted to prove. The definition of $F_\text{like}^{-1}(\cdot;\u)$ is readily verified by substitution and using the definition of $\lG_2$ (cf. \cref{def:invlt}). Indeed, \eqref{eq:invtri} implies that $\lF_2(\x,\lG_2(\lF_1(\x), \f)) = \f$ 
for every $\x$. Therefore, since the previous relation holds, in particular, for $\x = \lF_1^{-1}(\u)$ we have
    \begin{align*}
        F_\text{like}(\lG_2(\u,\f);\u) = \lF_2(\lF_1^{-1}(\u), \lG_2(\u,\f)) = \f
    \end{align*}
    as we wanted to prove.
\end{proof}
\begin{lemma}\label{lemma:post}
Let $\uF : \mathbb{R}^{n+m} \to \mathbb{R}^{n+m}$ be an invertible upper-triangular map with inverse $\uG$
satisfying
$$
\uF_\#\left(\mu_\X \otimes \mu_\Y\right) = \mu_{\U,\F},
$$
for given reference and target measures $\mu_\X, \mu_\Y$, $\mu_{\U,\F}$.
Then, for any $\f \in \mathbb{R}^m$, the map
$$
F_{\text{post}}(\cdot; \f) := \uF_1(\cdot,\uF_2^{-1}(\f)),
$$
with inverse
$$F_\text{post}^{-1}(\cdot;\f):=\uG_1(\cdot,\f),$$
pushes forward $\mu_\X$ to the conditional measure $\mu_{\U|\F=\f}$, i.e.,
$$
\mu_{\U|\F=\f} = F_{\text{post}}(\cdot; \f)_\# \mu_\X.
$$
\end{lemma}
\begin{proof}
    The proof is analogous to that of \cref{lemma:like}.
\end{proof}
A schematic depiction of the maps $F$, $F_\text{like}$ and $F_\text{post}$ is shown in \cref{fig:intro2}. It shows how, through a lower-triangular transport map between $\mu_{\X}\otimes\mu_{\Y}$ and $\mu_{\U,\F}$ we can condition on given $\u$ by first mapping $\u$ to latent space and subsequently pushing forward to obtain a sample from the posterior. We can sample from the likelihood in a similar fashion using an upper-triangular map. We refer to \cite{Kobyzev2021} for an overview of suitable triangular architectures which can be used to parametrize these invertible maps.

%%%%%%%%%%%%%%%%%%%%%%%%%%%%%%%%%%%%%%%%%%%%%%%%%%%%%%%%%%%%%%%%%%%%%%%%%%%%%%%%%%%%%%%%%%%%%%%%%%%%%%%%%%%
%%%%%%%%%%%%%%%%%%%%%%%%%%%%%%%%%%%%%%%%%%%%%%%%%%%%%%%%%%%%%%%%%%%%%%%%%%%%%%%%%%%%%%%%%%%%%%%%%%%%%%%%%%%
\section{A reversible simulator}\label{main}
%%%%%%%%%%%%%%%%%%%%%%%%%%%%%%%%%%%%%%%%%%%%%%%%%%%%%%%%%%%%%%%%%%%%%%%%%%%%%%%%%%%%%%%%%%%%%%%%%%%%%%%%%%%
%%%%%%%%%%%%%%%%%%%%%%%%%%%%%%%%%%%%%%%%%%%%%%%%%%%%%%%%%%%%%%%%%%%%%%%%%%%%%%%%%%%%%%%%%%%%%%%%%%%%%%%%%%%
While the previous section showed how we can construct two separate maps ($F_\text{like}$ and $F_\text{post}$) for simulation and inference, we now define an invertible mapping between $\mu_\U\otimes\mu_{\Y}$ and $\mu_\X\otimes\mu_{\F}$ with the desired property of allowing for both simulation and inference with a single invertible map. A schematic depiction of this map is shown in \cref{fig:intro}. We define it formally as follows.

\begin{definition}[Reversible Simulator]\label{def:RS}
An invertible map $S:\mathbb{R}^{n+m}\rightarrow\mathbb{R}^{n+m}$ with inverse $R :\mathbb{R}^{n+m}\rightarrow\mathbb{R}^{n+m}$ is a \emph{Reversible Simulator} for given reference and target measures $\mu_{\X},\mu_{\Y}$, and $\mu_{\U,\F}$ when it obeys the following properties
\begin{enumerate}
\item[i)] $\left(S_2(\u,\cdot)\right)_\#\mu_\Y = \mu_{\F|\U=\u},$
\item[ii)]$\left(R_1(\cdot,\f)\right)_\#\mu_\X = \mu_{\U|\F=\f},$
\item[iii)] $S_\#\left(\mu_{\U}\otimes\mu_{\Y}\right) = \left(\mu_{\X}\otimes\mu_{\F}\right),$
\item[iv)] $R_\# \left(\mu_{\X}\otimes\mu_{\F}\right) = \left(\mu_{\U}\otimes\mu_{\Y}\right),$
\end{enumerate}
where $S_1:\mathbb{R}^{n+m}\rightarrow\mathbb{R}^n$, $S_2:\mathbb{R}^{n+m}\rightarrow\mathbb{R}^m$ denote the components of $S = (S_1,S_2)$ and $(R_1,R_2)$ are similarly defined as the components of $R$.
\end{definition}
\begin{lemma}[Necessary conditions]\label{lemma:necessary}
Let $S$ be a Reversible Simulator and consider the properties outlined in \cref{def:RS}. Then
\begin{itemize}
\item i) and iii) imply ii).
\item ii) and iii) imply i),
\end{itemize}
\end{lemma}
\begin{proof} We will show that \emph{i)} and \emph{iii)} implies \emph{ii)}. The other statement follows similarly.
Given a test function $\varphi$, by \emph{i)} we have
\[
\mathbb{E}_{\mu_{\U}}\mathbb{E}_{\mu_{\F|\U}}\varphi(\u,\f) = \mathbb{E}_{\mu_{\U,\Y}}\varphi(\u,S_2(\u,\y)) = \mathbb{E}_{\mu_{\U,\Y}}\varphi(R_1\circ S(\u,\y),S_2(\u,\y)),
\]
where the second equality arises from invertibility of $S$. Then, by \emph{iii)} we have
\[
\mathbb{E}_{\mu_{\U,\Y}}\varphi(R_1\circ S(\u,\y),S_2(\u,\y)) = \mathbb{E}_{\mu_{\X,\F}}\varphi(R_1(\x,\f),\f)=\mathbb{E}_{\mu_\F}\mathbb{E}_{R_1(\cdot,\f)_\#\mu_\X}\varphi(\u,\f).
\]
By uniqueness of the disintegration kernel, this implies that $R_1(\cdot,\f)_\#\mu_\X = \mu_{\U|\F=\f}$, yielding te desired result.
\end{proof}
\begin{remark}
Note that \emph{iii)} and \emph{iv)} do not imply \emph{i)} and \emph{ii)} generally. Indeed, any map $S = (S_1,S_2)$ with $(S_1)_{\#}\mu_\X=\mu_\U$ and $(S_2)_{\#}\mu_\F=\mu_\Y$ will satisfy \emph{iii)} and \emph{iv)} without necessarily achieving \emph{i)} and \emph{ii)}.
\end{remark}
%%%%%%%%%%%%%%%%%%%%%%%%%%%%%%%%%%%%%%%%%%%%%%%%%%%%%%%%%%%%%%%%%%%%%%%%%%%%%%%%%%%%%%%%%%%%%%%%%%%%%%%%%%%
\subsection{Existence of Reversible Simulators}\label{sec:existence RS}
%%%%%%%%%%%%%%%%%%%%%%%%%%%%%%%%%%%%%%%%%%%%%%%%%%%%%%%%%%%%%%%%%%%%%%%%%%%%%%%%%%%%%%%%%%%%%%%%%%%%%%%%%%%
We can construct a reversible simulator from the triangular transport maps introduced in \cref{prelim}, as per the following lemma.
\begin{theorem}[Existence]\label{thm:mapsSR}
Let $F_{\text{post}}$, $F_\text{like}$ as defined in \cref{lemma:like,lemma:post} for reference and target measures $\mu_{\X},\mu_{\Y}$ and $\mu_{\U,\F}$. Define the map $S = (S_1,S_2)$ with inverse $R = (R_1,R_2)$ as
\begin{equation}\label{eq:mapS}
S_1(\u,\y) = F_\text{post}^{-1}(\u;F_\text{like}(\y;\u)), \quad S_2(\u,\y)=F_\text{like}(\y;\u),
\end{equation}
\begin{equation}\label{eq:mapR}
R_1(\x,\f) = F_\text{post}(\x;\f), \quad R_2(\x,\f) = F_\text{like}^{-1}(\f;F_\text{post}(\x;\f)),
\end{equation}
Then $S$ is a Reversible Simulator (cf. \cref{def:RS}) for the reference and target measures $\mu_{\X},\mu_{\Y}$, and $\mu_{\U,\Y}$.
\end{theorem}
\begin{proof}
    The relation $S^{-1} = R$ follows from a direct computation, while \emph{i)} and \emph{ii)} hold by construction (cf. \cref{lemma:like,lemma:post}). We now prove $iii)$ and $iv)$ and since $S^{-1} = R$, it is enough to prove $iii)$. Given a test function $\varphi$, it holds that 
    \begin{align*}
        &\mathbb{E}_{\mu_{\X,\F}} \varphi(R_1(\x,\f), R_2(\x,\f)) = \mathbb{E}_{\mu_{\X,\F}} \varphi(F_{\rm post}(\x;\f), F_\text{like}^{-1}(\f;F_\text{post}(\x;\f))) = \\
        &\mathbb{E}_{\mu_{\X,\F}} \varphi(\uF_1(\x, \uF_2^{-1}(\f)), F_\text{like}^{-1}(\f;\uF_1(\x, \uF_2^{-1}(\f))))
             = \mathbb{E}_{\mu_{\X,\Y}} \varphi(\uF_1(\x, \y), F_\text{like}^{-1}(\uF_2(\y);\uF_1(\x, \y)))
    \end{align*}
    where we used the definition of $F_{\text{post}}$ in \cref{lemma:post} and that $(\uF_2^{-1})_\#\mu_\F = \mu_\Y$.
    Now, using the definition of $F_\text{like}^{-1}$ (cf. \cref{lemma:like}) we can continue the previous estimate as
    \begin{align*}     
     & = \mathbb{E}_{\mu_{\X,\Y}} \varphi(\uF_1(\x, \y), \lF_2^{-1}(\uF_1(\x,\y),\uF_2(\y)) )
        = \mathbb{E}_{\mu_{\U,\F}} \varphi(\u, \lG_2(\u,\f))\\
     & = \mathbb{E}_{\mu_{\U,\F}} \varphi(\lF_1(\lF^{-1}_1(\u)), \lG_2(\u,\f))
        = \mathbb{E}_{\mu_{\X,\Y}} \varphi(\lF_1(\x), \y)
             = \mathbb{E}_{\mu_{\U,\Y}} \varphi(\u, \y)
\end{align*}
where additionally we used that $\lF^{-1}_\# \mu_{\U , \F} = \mu_{\X}\otimes\mu_{\Y}$ and that $(\lF_1)_\# \mu_\X = \mu_\U$. Because the result holds for arbitrary $\varphi$, this proves $(iii)$.
\end{proof}
\begin{corollary}[Decomposition of $S$]
We can alternatively express the map $S$ constructed in \cref{thm:mapsSR} in terms of $\lF$, $\uF$ and their inverse $\lG, \uG$ (cf. \cref{lemma:like,lemma:post}) as
\[
S = U \circ L, \quad L(\u,\y)=\left(\u,\lF_2(\lG_1(\u),\y)\right), U(\u,\f)=\left(\uG_1(\u,\f),\f\right).
\]
\end{corollary}
\begin{proof}
By construction we then have
\[
S_2(\cdot;\u)_\#\pi_\Y = \mu_{\F|\U=\u}, \quad R_1(\cdot;\f)_\#\pi_\X = \mu_{\U|\F=\f},
\]
such that $\f = S_2(\u,\y)$ with $\y\sim\mu_\Y$ produces a sample from the likelihood (i.e., a simulation) and $\u = R_1(\x,\f)$ with $\x\sim\mu_\X$ produces a sample from the posterior (i.e., inference). 
\end{proof}
% The Reversible Simulator constructed through triangular maps in addition has the following useful property.
% \begin{corollary}\label{lemma:mapsSR2}
% Let the maps $S$ and $R$ as defined in \eqref{eq:mapS}, \eqref{eq:mapR}. Then, it holds that 
% \begin{itemize}
% \item [i)] $\mathbb{E}_{\mu_\X}[(R_2(\x,\cdot))_\# \mu_\F] = \mu_\Y$.
% \item [ii)]  $\mathbb{E}_{\mu_\Y}[(S_1(\cdot,\y))_\#\mu_\U] = \mu_\X$.
% \end{itemize}  
% \end{corollary}
% \begin{proof}
% We again prove only the statement $i)$. Given an arbitrary test function $\varphi$ it holds that 
% \begin{align*}
% \mathbb{E}_{\mu_\X}\mathbb{E}_{\mu_\F} \varphi(R_2(\f,\x)) &  =  \mathbb{E}_{\mu_\X}\mathbb{E}_{\mu_\F} \varphi(F_\text{like}^{-1}(\f;F_\text{post}(\x;\f))) \\
% & = \mathbb{E}_{\mu_\X}\mathbb{E}_{\mu_\F} \varphi( F_\text{like}^{-1}(\f;\uF_1(\x,\uF_2^{-1}(\f)))\\
% & =  \mathbb{E}_{\mu_\X}\mathbb{E}_{\mu_\F} \varphi( \lF_2^{-1}(\uF_1(\x,\uF_2^{-1}(\f)),\f))\\
% & = \mathbb{E}_{\mu_\X}\mathbb{E}_{\mu_\Y} \varphi( \lF_2^{-1}(\uF_1(\x,\y),\uF_2(\y)))\\
% & = \mathbb{E}_{\mu_{\U,\F}} \varphi(\lF_2^{-1}(\u,\f))\\
% & = \mathbb{E}_{\mu_\X}\mathbb{E}_{\mu_\Y} \varphi(\y)\\
% & = \mathbb{E}_{\mu_\Y} \varphi(\y)
% \end{align*}
% as we wanted to prove.
% \end{proof}
%%%%%%%%%%%%%%%%%%%%%%%%%%%%%%%%%%%%%%%%%%%%%%%%%%%%%%%%%%%%%%%%%%%%%%%%%%%%%%%%%%%%%%%%%%%%%%%%%%%%%%%%%%%
\subsection{Uniqueness}
%%%%%%%%%%%%%%%%%%%%%%%%%%%%%%%%%%%%%%%%%%%%%%%%%%%%%%%%%%%%%%%%%%%%%%%%%%%%%%%%%%%%%%%%%%%%%%%%%%%%%%%%%%%
The Reversible Simulator as constructed through lower and upper triangular maps (cf. \cref{thm:mapsSR}) is not unique. Given \emph{a} Reversible Simulator, we can construct another one by left and right composition with invertible maps that preserve the reference densities. We make this precise in the following statement.

\begin{lemma}[Uniqueness] Let $S:\mathbb{R}^{n+m}\rightarrow \mathbb{R}^{n+m}$ be a Reversible Simulator with inverse $R$ for reference and target measures $\mu_\X \otimes \mu_\Y$, $\mu_{\U,\F}$ (cf. \cref{def:RS}) and let $P=(P_1, I_m)$ and $Q=(I_n,Q_2)$ be invertible measure-preserving maps in the sense that
\[
P_1(\cdot,\f)_\# \mu_\X = \mu_\X, \quad Q_2(\u,\cdot)_\#\mu_\Y = \mu_\Y \ \ \text{for all } \  \f,\u,
\]
then
\[
S' = P^{-1} \circ S \circ Q, 
\]
with inverse $R' = Q^{-1} \circ R \circ P$,
is also a Reversible Simulator.
\end{lemma}
\begin{proof}
We show that the with the stated maps $P,Q$, the reversible simulator $S' = P^{-1}\circ\overline{S}\circ Q$ satisfies the conditions of \cref{def:RS}, given that $S$ satisfies them. First, note that 
\[
P_\# \left(\mu_{\X}\otimes\mu_{\F}\right) = \left(\mu_{\X}\otimes\mu_{\F}\right), \quad Q_\# \left(\mu_{\U}\otimes\mu_{\Y}\right)=\mu_{\U}\otimes\mu_{\Y}.
\]
Thus conditions \emph{iii)} and \emph{iv)} are satisfied. Furthermore, note that
\[
S'_2(\u,\y) = S_2(\u,Q_2(\u,\y)), \quad R'_1(\x,\f) = R_1(P_1(\x,\f),\f).
\]
Thus,
\[
S'_2(\u,\cdot)_\#\mu_\Y = S_2(\u,\cdot)_\#\left(Q_2(\u,\cdot)_\#\pi_\Y\right) = S_2(\u,\cdot)_\#\mu_\Y \equiv \mu_{\F|\U=\u},
\]
and similarly
\[
R'_1(\cdot,\f)_\#\mu_\X = R_1(\cdot,\f)_\#\left(P_1(\cdot,\f)_\#\mu_\X\right)= \R_1(\cdot,\f)_\#\mu_\X \equiv \mu_{\U|\F=\f},
\]
verifying that $S'$ satisfies conditions \emph{i)} and \emph{ii)}. 
\end{proof}

\subsection{A variational formulation}
%%%%%%%%%%%%%%%%%%%%%%%%%%%%%%%%%%%%%%%%%%%%%%%%%%%%%%%%%%%%%%%%%%%%%%%%%%%%%%%%%%%%%%%%%%%%%%%%%%%%%%%%%%%
While we can now construct $S$ explicitly from the two aforementioned triangular maps (cf. \cref{thm:mapsSR}), we would like to be able to use generic invertible network architectures to represent $S$. Therefore, we present the following result which entails a training loss that can be evaluated directly on $S$ and $S^{-1}$.

\begin{lemma}[Variational formulation]\label{cor:J1J2}
Given $S = (S_1, S_2)$ with $S^{-1} = R = (R_1,R_2)$ and reference and target measures $\mu_{\X},\mu_{\Y}$ and $\mu_{\U,\F}$, consider the following losses:
\begin{align}
\mathcal{J}_1(S) &= \mathcal{M}\bigl(\left(I_n,  S_2\right)_\#\left(\mu_{\U}\otimes\mu_{\Y}\right), \mu_{\U,\F}\bigr),\\
\mathcal{J}_2(R) &= \mathcal{M}\bigl(\left(R_1, I_m\right)_\#\left(\mu_{\X}\otimes\mu_{\F}\right), \mu_{\U,\F}\bigr),\\
\mathcal{J}_3(S) &= \mathcal{M}\bigl( S_\# \left(\mu_{\U}\otimes\mu_{\Y}\right), \left(\mu_{\X}\otimes\mu_{\F}\right)\bigr),\\
\mathcal{J}_4(R) &= \mathcal{M}\bigl( R_\# \left(\mu_{\X}\otimes\mu_{\F}\right), \left(\mu_{\U}\otimes\mu_{\Y}\right)\bigr),
\end{align}
with
$$\left(I_n, S_2\right)(\u,\y) = \left(\u,S_2(\u,\y)\right), \left(R_1, I_m\right)(\x,\f) = \left(R_1(\x,\f), \f\right),$$
and where $\mathcal{M}$ is some distance on the space of measures. Then the following statements hold:
\begin{enumerate}
\item[i)] $\mathcal{J}_1(S) = 0$ if and only if $(S_2(\u,\cdot))_\#\pi_\Y = \mu_{\F|\U=\u},$
\item[ii)] $\mathcal{J}_2(R) = 0$  if and only if $\left(R_1(\cdot,\f)\right)_\#\pi_\X = \mu_{\U|\F=\f},$
\item[iii)] $\mathcal{J}_3(S) = 0$ if and only if $\mathcal{J}_4(R) = 0,$
\item [iv)]  $\mathcal{J}_3(S) = \mathcal{J}_1(S) = 0$ implies  $\mathcal{J}_2(R) = 0,$
\item [v)] $\mathcal{J}_2(R) = \mathcal{J}_4(R) = 0$ implies  $\mathcal{J}_1(S) = 0$.
\end{enumerate}
\end{lemma}
\begin{proof}
Suppose that $\mathcal{J}_1(S) = 0$. Since $\mathcal{M}$ is a distance on the space of measures it holds that $\mu_\U \otimes [(S_2(\u,\cdot))_\#\mu_\Y] = \mu_{\U,\F}$. Therefore, the relation $(S_2(\u,\cdot))_\#\pi_\Y = \mu_{\F|\U=\u}$ follows immediately from the uniqueness of conditional expectation. Vice versa, if  $(S_2(\u,\cdot))_\#\pi_\Y = \mu_{\F|\U=\u}$, then, it also holds that 
$\mu_\U \otimes [(S_2(\u,\cdot))_\#\mu_\Y] = \mu_{\U,\F}$ implying that $\mathcal{J}_1(S) = 0$. The proof of $ii)$ is similar.
Note that $iii)$ is immediate from the definition of the push-forward and the assumption $R = S^{-1}$. We now prove $iv)$ (since proof of $v)$ is similar we omit it). Since $\mathcal{J}_1(S) = 0$ implies that $(S_2(\u,\cdot))_\#\mu_\Y = \mu_{\F|\U=\u}$. Therefore, for every test function $\varphi$ it holds 
\begin{align*}
&\mathbb{E}_{\mu_{\U, \F}} \varphi(\u,\f) = \mathbb{E}_{\mu_{\U}\otimes\mu_{\Y}} \varphi(\u, S_2(\y,\u))  =\mathbb{E}_{\mu_{\U}\otimes\mu_{\Y}} \varphi(R_1(S(\y,\u)), S_2(\y,\u)) = \\
&\mathbb{E}_{\mu_{\F}\otimes\mu_{\X}} \varphi(R_1(\x,\f), S_2(R(\x,\f))) 
= \mathbb{E}_{\mu_{\F}\otimes\mu_{\X}} \varphi(R_1(\x,\f), \f)
=\mathbb{E}_{\mu_\F} \mathbb{E}_{R_1(\cdot,\f)_\#\mu_\X} \varphi(\u,\f), 
\end{align*}
where in the second equality we used that $R = S^{-1}$ and in the third equality we used that $S_\#  \pi_{\U \times \Y} = \pi_{\F\times \X}$ that is a consequence of $\mathcal{J}_3(S) = 0$. We thus conclude that $\left(R_1(\cdot,\f)\right)_\#\pi_\X = \mu_{\U|\F=\f}$ and therefore $\mathcal{J}_2(R) = 0$.
\end{proof}

For a specific choice of $\mathcal{M}$ it is possible to obtain quantitative estimates for the previous lemma. Below we show that this is the case for $\mathcal{M}$ being the $p$-Wasserstein distance with $1\leq p <\infty$. 
\begin{corollary}[Approximation]
    Let $\mathcal{M} = W_p$ with $1\leq p<\infty$. Suppose that $S$ is bi-Lipschitz continuous with constant $\alpha$, then
\begin{align}\label{eq:appr1}
\mathcal{J}_3(S) \leq \varepsilon \Rightarrow \mathcal{J}_4(R) \leq \alpha \varepsilon
\end{align}
\begin{align}\label{eq:appr2}
\mathcal{J}_4(R) \leq \varepsilon \Rightarrow \mathcal{J}_3(S) \leq \alpha \varepsilon.
\end{align}    
 Moreover, 
    \begin{align}\label{eq:appr3}
    \mathcal{J}_1(S) + \mathcal{J}_3(S) \leq \varepsilon \Rightarrow \mathcal{J}_2(R) \leq\sqrt{1 + \alpha^2} \varepsilon
    \end{align}
    and
 \begin{align}\label{eq:appr4}
   \mathcal{J}_2(R) + \mathcal{J}_4(R) \leq \varepsilon \Rightarrow \mathcal{J}_1(S) \leq\sqrt{1 + \alpha^2} \varepsilon.
 \end{align}
\end{corollary}
\begin{proof}
The proof is based on the following property of the Wasserstein distances:
\begin{align}\label{eq:Lipest}
    W_p(F_\#\mu , F_\#\nu) \leq {\rm Lip}(F)W_p(\mu,\nu)
\end{align}
for $\mu,\nu$ given probability measures and $F$ any Lipschitz map. The proof of such property is standard and can be found, for example, in \cite[Lemma C.1]{patrini2020sinkhorn}. 

We first prove \eqref{eq:appr1}. Note that since $S^{-1} = R$ and using \eqref{eq:Lipest} we have that
\begin{align*}
\mathcal{J}_4(R) &= W_p\bigr(R_\# \left(\mu_{\X}\otimes\mu_{\F}\right), \left(\mu_{\U}\otimes\mu_{\Y}\right)\bigr) = W_p\bigr(R_\# \left(\mu_{\X}\otimes\mu_{\F}\right), R_\#S_\#\left(\mu_{\U}\otimes\mu_{\Y}\right)\bigr) \\
& \leq {\rm Lip}(R) W_p\bigr(\left(\mu_{\X}\otimes\mu_{\F}\right),S_\#\left(\mu_{\U}\otimes\mu_{\Y}\right)\bigr) = {\rm Lip}(R) \mathcal{J}_3(S) \leq \alpha\varepsilon 
\end{align*}
as we wanted to prove. The statement \eqref{eq:appr2} follows similarly.

We now prove \eqref{eq:appr3}. First we compute the Lipschitz constant of the map $(\x,\f)\mapsto (R_1, I_m)(\x,\f)$. Note that 
\begin{align*}
\|(R_1, I_m)(\x,\f)& - (R_1, I_m)(\x',\f')\|^2_2 = \|R_1(\x,\f) - R_1(\x',\f')\|^2_2 + \|\f_1 - \f_2\|_2^2 \\
& \leq {\rm Lip}(R_1)^2 \|(\x,\f) - (\x',\f')\|_2^2 + \|\f - \f'\|_2^2 \leq (1 + {\rm Lip}(R_1)^2) \|(\x,\f) - (\x',\f')\|_2^2, 
\end{align*}
proving that the Lipschitz constant of $(R_1, I_m)$ is bounded from above by $\sqrt{1 + {\rm Lip}(R_1)^2}$. Moreover, by Lipschitz continuity of $R$ with constant $\alpha$ it follows that $R_1(\u,\cdot)$ is Lipschitz continuous with constant (at most) $\alpha$ as well. Thus the Lipschitz constant of the map $(R_1, I_m)$ is bounded by $\sqrt{1+\alpha^2}$.

Now using the triangle inequality for the Wasserstein distance, \eqref{eq:Lipest} and the previous estimate of the Lipschitz constant we have that 
\begin{align*}
\mathcal{J}_2(R)\! & =\! W_p\bigl(\left(R_1, I_m\right)_\#\left(\mu_{\X}\otimes\mu_{\F}\right), \mu_{\U,\F}\bigr) \\
& \leq W_p\bigl(\left(R_1, I_m\right)_\#\!\left(\mu_{\X}\otimes\mu_{\F}\right)\!,\! \left(R_1, I_m\right)_\#\!S_\#\!(\mu_{\U} \otimes \mu_{\Y}) \bigr)\!\! + \!\! W_p\bigl(\left(R_1, I_m\right)_\#\!S_\#\!(\mu_{\U} \otimes \mu_{\Y})\!,\!\mu_{\U,\F}\bigr)\\
& \leq \sqrt{1 + {\rm Lip}(R_1)^2}  W_p\bigl(\left(\mu_{\X}\otimes\mu_{\F}\right), S_\#(\mu_{\U} \otimes \mu_{\Y}) \bigr) + \mathcal{J}_1(S) \\
& = \sqrt{1 + {\rm Lip}(R_1)^2}  \mathcal{J}_3(S) + \mathcal{J}_1(S)  \leq \sqrt{1 + \alpha^2} \varepsilon
\end{align*}
as we wanted to prove. \eqref{eq:appr4} can be proven similarly.
\end{proof}
\begin{remark}
While there is no guarantee that the exact map $S$ is bi-Lipschitz, it can be forced by the architecture or by adding a regularization term of the form $\log|\nabla S| + \log|\nabla R|$.
\end{remark}
%%%%%%%%%%%%%%%%%%%%%%%%%%%%%%%%%%%%%%%%%%%%%%%%%%%%%%%%%%%%%%%%%%%%%%%%%%%%%%%%%%%%%%%%%%%%%%%%%%%%%%%%%%%
%%%%%%%%%%%%%%%%%%%%%%%%%%%%%%%%%%%%%%%%%%%%%%%%%%%%%%%%%%%%%%%%%%%%%%%%%%%%%%%%%%%%%%%%%%%%%%%%%%%%%%%%%%%
\section{Examples}\label{numerics}
%%%%%%%%%%%%%%%%%%%%%%%%%%%%%%%%%%%%%%%%%%%%%%%%%%%%%%%%%%%%%%%%%%%%%%%%%%%%%%%%%%%%%%%%%%%%%%%%%%%%%%%%%%%
%%%%%%%%%%%%%%%%%%%%%%%%%%%%%%%%%%%%%%%%%%%%%%%%%%%%%%%%%%%%%%%%%%%%%%%%%%%%%%%%%%%%%%%%%%%%%%%%%%%%%%%%%%%

%%%%%%%%%%%%%%%%%%%%%%%%%%%%%%%%%%%%%%%%%%%%%%%%%%%%%%%%%%%%%%%%%%%%%%%%%%%%%%%%%%%%%%%%%%%%%%%%%%%%%%%%%%%
\subsection{A linear inverse problem with Gaussian distributions}\label{linearexample}
%%%%%%%%%%%%%%%%%%%%%%%%%%%%%%%%%%%%%%%%%%%%%%%%%%%%%%%%%%%%%%%%%%%%%%%%%%%%%%%%%%%%%%%%%%%%%%%%%%%%%%%%%%%
Consider a linear inverse problem with a Gaussian prior $N(\mathbf{0}, \Sigma_\U)$ and likelihood $N(K\u, \Sigma_\F)$, with $K\in\mathbb{R}^{m\times n}$ a given matrix. The joint density is Gaussian with mean zero and covariance
\[
\Sigma_{\U,\F} = \left(\begin{matrix} \Sigma_\U & \Sigma_\U K^\top \\ K\Sigma_\U & K\Sigma_\U K^\top + \Sigma_\F \end{matrix}\right).
\]
The conditional distributions are the likelihood with mean $K\u$ and variance $\Sigma_\text{like} = \Sigma_\F $ and the posterior with mean $(K^\top\Sigma_\F^{-1}K + \Sigma_\U^{-1})^{-1}K^\top \Sigma_\F^{-1}\f$ and covariance $\Sigma_\text{post} = (K^\top\Sigma_\F^{-1}K + \Sigma_\U^{-1})^{-1}$.

In the case of normalizing flows, we look for $F$ such that $\mu_{\U, \F} = F_\# (\mu_{\X}\otimes\mu_{\Y})$. In this case a linear map suffices and we get the following matrix-representations of the lower and upper triangular maps
\[
\widecheck{F} = \left(\begin{matrix} \Sigma_\U^{1/2} & 0 \\ K\Sigma_\U^{1/2} & \Sigma_\F^{1/2}\end{matrix}\right),\qquad
\widehat{F} = \left(\begin{matrix} \Sigma_\text{post}^{1/2} & \Sigma_\U K^\top (K\Sigma_\U K^\top + \Sigma_\U)^{-1/2}\\ 0 & (K\Sigma_\U K^\top + \Sigma_\F)^{1/2}\end{matrix}\right),
\]
where $A^{1/2}$ denotes a matrix for which $(A^{1/2})(A^{1/2})^\top = A$ and depending on the context is either a lower or upper triangular Cholesky decomposition.

The construction proposed in \eqref{eq:mapS} and \eqref{eq:mapR} yields
\begin{equation}\label{eq:gaussianS}
S = \left(\begin{matrix}\Sigma_\text{post}^{1/2}\Sigma_\U^{-1} & -\Sigma_\text{post}^{1/2}K^\top\Sigma_\F^{-1/2} \\ K & \Sigma_{\F}^{1/2}\end{matrix}\right),
\end{equation}
with inverse
\[R = \left(\begin{matrix}\Sigma_\text{post}^{1/2} & \Sigma_\text{post}K^\top\Sigma_\F^{-1} \\ -\Sigma_{\F}^{-1/2}K\Sigma_\text{post}^{1/2} & \Sigma_{\F}^{1/2}\left(\Sigma_\F + K\Sigma_\U K^\top\right)^{-1}\end{matrix}\right).
\]

It is well-known that the triangular transport maps can be very ill-conditioned, and this is also observed here. We see that $\lF$ for example degenerates when $\|\Sigma_\F\|\rightarrow 0$ (i.e., a deterministic forward model). Note also that the same holds for $\uF$, as both $\lF$ and $\lF$ are square roots of $\Sigma_{\U,\F}$ and hence have the same condition number. The map $S$ on the other hand, does not degenerate as $\|\Sigma_\F\|\rightarrow 0$. If $m=n$ and $K$ is invertible, for example, we have 
$$S = \left(\begin{matrix}0_{n\times  n} & -I_{n\times n}\\ K & 0_{n\times n}\end{matrix}\right),$$
with $\kappa(S) = \|K\|_2$. The reversible simulator is well-defined for a deterministic forward model and in that case reflects the ill-posedness of the forward model (as it should because it simulates the data).
%%%%%%%%%%%%%%%%%%%%%%%%%%%%%%%%%%%%%%%%%%%%%%%%%%%%%%%%%%%%%%%%%%%%%%%%%%%%%%%%%%%%%%%%%%%%%%%%%%%%%%%%%%%
\subsection{A non-linear example}\label{nonlinearexample}
%%%%%%%%%%%%%%%%%%%%%%%%%%%%%%%%%%%%%%%%%%%%%%%%%%%%%%%%%%%%%%%%%%%%%%%%%%%%%%%%%%%%%%%%%%%%%%%%%%%%%%%%%%%
Consider the lower triangular map $\widecheck{F}: \mathbb{R}^2 \rightarrow \mathbb{R}^2$ defined by 
$$\widehat F(x, y) = \begin{pmatrix}
    ax \\ \text{sign}(x) + by
\end{pmatrix},$$
for $(x, y)\in \mathbb{R}^2$ and fixed $a, b>0$. Let $\mu_{\X, \Y} = N(0, I_2)$ and set $\mu_{\U, \F} = \widecheck F_\# \mu_{\X, \Y}$. Note that $\widecheck F$ is invertible and is the lower triangular Knothe-Rosenblatt rearrangement from $\mu_{\X, \Y}$ to $\mu_{\U, \F}$. Let $\Phi$ denote the cumulative distribution function of a standard normal distribution on $\mathbb{R}$ and define $\psi(f) := \frac{1}{1 + \exp(-2f / b^2)}$ for $f\in\mathbb{R}$. An elementary calculation shows that the upper triangular Knothe-Rosenblatt rearrangement $\widehat F: \mathbb{R}^2\rightarrow\mathbb{R}^2$ is given by $\widehat F(x, y) = (\widehat F_1(x, y), \widehat F_2(y))$ with $\widehat F_1$ given by
\begin{equation*}
    \widehat F_1(x, y) = \begin{cases}
        a \Phi^{-1}\left(\frac{\Phi(x)}{2(1-\psi(\widehat F_2(y)))}\right) & \text{if} \quad \Phi(x) < 1 - \psi(\widehat F_2(y))\\
        a \Phi^{-1}\left(\frac{\Phi(x) - 1 + 2\psi(\widehat F_2(y))}{2\psi(\widehat F_2(y))}\right) & \text{else}
    \end{cases}
\end{equation*}
and $\widehat F_2 = C_F^{-1}\circ \Phi$, where $C_F$ is the cumulative distribution function of the second marginal of $\mu_{\U, \F}$ that is given by 
$$C_F(f) = \frac{1}{2}\left[\Phi\left(\frac{f+1}{b}\right) + \Phi\left(\frac{f-1}{b}\right)\right].$$
%The details can be found in the appendix.

Using the construction from \cref{thm:mapsSR}, a reversible simulator can be constructed that can be written as $S = U\circ L$ for 
$$U(u, f) = \begin{pmatrix}
    F_{\text{post}}^{-1}(u; f)\\ f
\end{pmatrix}\quad \text{and}\quad L(u, y) = \begin{pmatrix}
    u \\ \text{sign}(u) + by
\end{pmatrix},$$
where 
$$F_{\text{post}}^{-1}(u; f) = \Phi^{-1}(C_{U\mid F=f}(u)) = \begin{cases}
    \Phi^{-1}\left[1 - 2\psi(f) + 2\psi(f)\Phi(u/a)\right] & \text{if} \quad u\geq 0\\
    \Phi^{-1}\left[2(1-\psi(f))\Phi(u/a)\right] & \text{else}
\end{cases}.$$

\begin{figure}[H]
    \centering
    \includegraphics[width=0.7\textwidth]{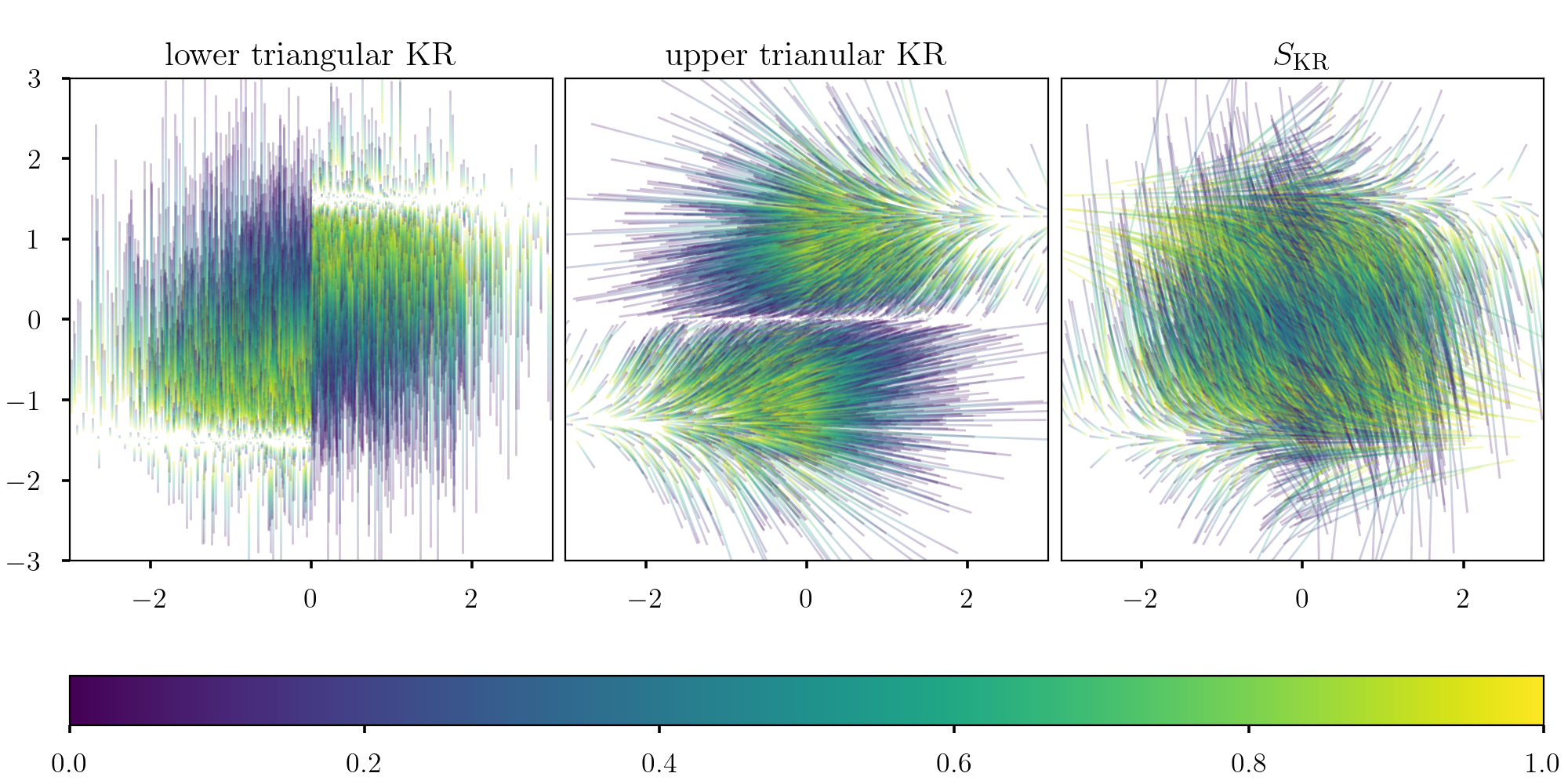}
    \caption{A flow map is plotted of the KR rearrangement, upper triangular KR rearrangement and the reversible simulator for $a=1$ and $b = 1/3$. Each line segment corresponds to a linear interpolation between a randomly generated point $z$ and its image under the transport map.}
    \label[figure]{fig:non_lin_example KR maps}
\end{figure}

In \cref{fig:non_lin_example KR maps}, the lower and upper triangular KR maps are visualized along with the reversible simulator constructed from them for $a=1$ and $b = \frac{1}{3}$. It shows that the different maps lead to very different transport plans.

% It is also possible to directly learn a reversible simulator using the losses $\mathcal{J}_1,\dots, \mathcal{J}_4$. The reversible simulator is parametrized by a class of affine coupling layers, where the scaling and translation functions are parametrized by standard feedforward neural networks. For the losses $\mathcal{J}_1, \mathcal{J}_2$ and $\mathcal{J}_4$ the maximum mean discrepancy is used. For $\mathcal{J}_3$, we note that since $\mu_{\U}\otimes \mu_{\Y}$ is Gaussian, it is possible to the KL-divergence as the pushforward density is explicitly known. In \cref{fig:non lin example learned S flow} a flow map is shown for the learned RS. Note the similarity between the learned RS and the Knothe-Rosenblatt RS from \cref{fig:non_lin_example KR maps}.

% \begin{figure}[H]
%     \centering
%     \includegraphics[width=0.4\textwidth]{figures/Nonlinear2.png}
%     \caption{A flow map is plotted of learned reversible simulator for $a=1$ and $b = 1/3$. Each line segment corresponds to a linear interpolation between a randomly generated point $z$ and its image $S(z)$.}
%     \label[figure]{fig:non lin example learned S flow}
% \end{figure}

%%%%%%%%%%%%%%%%%%%%%%%%%%%%%%%%%%%%%%%%%%%%%%%%%%%%%%%%%%%%%%%%%%%%%%%%%%%%%%%%%%%%%%%%%%%%%%%%%%%%%%%%%%%
\subsection{Inpainting}\label{inpainting}
%%%%%%%%%%%%%%%%%%%%%%%%%%%%%%%%%%%%%%%%%%%%%%%%%%%%%%%%%%%%%%%%%%%%%%%%%%%%%%%%%%%%%%%%%%%%%%%%%%%%%%%%%%%
Consider the problem of image inpainting. In this example we use the MNIST dataset consisting of $70000$ handwritten digits divided in $60000$ for training and $10000$ for testing. The images are grayscale with integer values from $0$ to $255$, which we scaled to floating-point values in $[0, 1]$. For each image we remove a number of rows from the center of the image and we add Gaussian noise. The forward operator is hence linear and can be described as
$\f = K\u + \boldsymbol{\eta}$, 
where $\u\in\R^{1024}$ is a flatted $32$ by $32$ image, $K \in \mathbb{R}^{1024 \times 1024}$ is the identity matrix with zeros on the rows of the pixels which we want to remove. The noise $\boldsymbol{\eta}$ is i.i.d. Gaussian with mean zero and variance $\sigma=0.1$.

% Taking a Gaussian prior $N(0, \Sigma)$ on $\u$ yields a jointly Gaussian distribution $\mu_{\u, \f}$. Therefore a lower or upper triangular square root of its covariance matrix yields a lower or upper triangular map $\widecheck F$ or $\widehat F$, respectively. Using the construction from \cref{sec:existence RS} it is then possible to directly calculate a reversible simulator, which is a linear map in this setting. In \cref{fig:posterior samples image inpainting} several samples are shown from the posterior distribution, where $\Sigma$ has been estimated using the standard maximum likelihood estimator from the samples. 

% \begin{figure}[!ht]
%     \centering
%     \includegraphics[width=0.8\textwidth]{figures/inpainting/linear_posterior_samples.png}
%     \caption{Using the relation $\mu_{\U\mid \F=\f} = R_1(\cdot , \f)_\# \mu_X$, nine posterior samples are generated and shown of the posterior distribution with $\F=\f_9$ shown in the top left.}
%     \label[figure]{fig:posterior samples image inpainting}
% \end{figure}

% \begin{figure}[!ht]
%     \centering
%     \includegraphics[width=0.5\textwidth]{figures/inpainting/linear_posterior_mean_var.png}
%     \caption{The posterior mean and pixel-wise variance of $\mu_{\U\mid \F=\f_9}$ are plotted with $\f_9$ as in the top left of \cref{fig:posterior samples image inpainting}.}
%     \label[figure]{fig:posterior mean var image inpainting}
% \end{figure}

\subsubsection{Implementation}
Using the samples $(\u,\f)$ generated as indicated above, we learn a reversible simulator using the losses $\mathcal{J}_1,\dots, \mathcal{J}_4$ (cf. \cref{cor:J1J2}). The noise distributions $\mu_\X$ and $\mu_\Y$ are chosen as standard normal Gaussians with dimension $32\times 32$. The vector $(\u,\y)$ consisting of an image and noise, is shaped as a single $32$ by $32$ image with two channels; one grayscale image channel and one noise channel. We treat the vector $(\x,\f)$ in a similar fashion.

To parametrize $S$, we use an adaptation of the GLOW network architecture introduced in \cite{NEURIPS2018_d139db6a} (see also \cite{dinh2017density}). In general the architecture remains entirely the same, with the exception that we remove the split layer and instead have both an image and noise going into the network and going out of the network. An overview is given in \cref{fig:flow_architecture}. For the split in the affine coupling layers, we only split along the channel dimension. However, in order to still mix the noise with the image, the squeeze layer is used. It splits the image into $(2, 2, c)$ blocks and rearranges them into $(1, 1, 4c)$ blocks, increasing the number of channels while shrinking the image by a factor of $2$. For more details on the GLOW network architecture, we refer the reader to \cite{NEURIPS2018_d139db6a}.

\begin{figure}[htbp]
    \centering
    \begin{tikzpicture}[
        font=\footnotesize, 
        >=Triangle,
        arrow/.style={->, thick, >=Triangle, line width=0.8pt},
        std_box/.style={
            draw=bordergray, 
            thick, 
            minimum width=3.0cm, 
            minimum height=0.55cm, 
            align=center
        },
        step_box/.style={std_box, top color=white, bottom color=gray!20},
        flow_box/.style={std_box, fill=boxgreen},
        squeeze_box/.style={std_box, fill=boxyellow},
        var_node/.style={
            draw=bordergray, 
            thick, 
            fill=gray!10, 
            inner sep=3pt, 
            rounded corners=0.25cm,
            minimum height=0.55cm,
            minimum width=0.55cm
        }
    ]

    % ==========================================
    % (a) One step of our flow
    % ==========================================
    \begin{scope}[local bounding box=partA]
        \node[step_box] (actnorm) at (0, 1.2) {actnorm};
        \node[step_box] (conv1x1) [above=0.35cm of actnorm] {invertible 1x1 conv};
        \node[step_box] (affine) [above=0.35cm of conv1x1] {affine coupling layer};

        \begin{scope}[on background layer]
            \node[fit=(actnorm) (affine), draw=bordergray, thick, fill=bggreen, inner xsep=0.4cm, inner ysep=0.5cm] (bg_box) {};
        \end{scope}

        \coordinate (in_A) at ([yshift=-0.7cm]actnorm.south);
        \coordinate (out_A) at ([yshift=0.7cm]affine.north);
        
        \draw[arrow] (in_A) -- (actnorm.south);
        \draw[arrow] (actnorm.north) -- (conv1x1.south);
        \draw[arrow] (conv1x1.north) -- (affine.south);
        \draw[arrow] (affine.north) -- (out_A);
    \end{scope}

    % ==========================================
    % (b) Multi-scale architecture
    % ==========================================
    \begin{scope}[xshift=6.2cm, local bounding box=partB]
        
        \node[squeeze_box] (sq1) at (0, 0.5) {squeeze};
        \node[var_node] (input) [below=0.4cm of sq1] {$(\u, \y)$};
        
        \node[flow_box] (flow1) [above=0.35cm of sq1] {step of flow};
        \node[right=0.15cm of flow1] {$\times$ $k$};

        \node[squeeze_box] (sq2) [above=0.6cm of flow1] {squeeze};
        \node[flow_box] (flow2) [above=0.35cm of sq2] {step of flow};
        \node[right=0.15cm of flow2] {$\times$ $k$};
        \node[var_node] (output) [above=0.4cm of flow2] {$(\x, \f)$};

        \draw[arrow] (input.north) -- (sq1.south);
        \draw[arrow] (sq1.north) -- (flow1.south);
        \draw[arrow] (flow1.north) -- (sq2.south);
        \draw[arrow] (sq2.north) -- (flow2.south);
        \draw[arrow] (flow2.north) -- (output.south);

        \coordinate (branch_start) at ($(flow1.north)!0.5!(sq2.south)$);
        \coordinate (return_height) at ($(input.north)!0.5!(sq1.south)$);
        
        % Fixed: Increased horizontal extension from 2.2cm to 2.5cm to completely clear the "x K" label
        \draw[arrow] (branch_start) 
            -- ++(2.5cm, 0) coordinate (top_right)
            -- (top_right |- return_height) coordinate (bottom_right) node[midway, right] {$\times$ ($\ell-1$)}
            -- (return_height);

    \end{scope}

    % ==========================================
    % Labels
    % ==========================================
    \node (labelA) at (0, -1.3) {(a) One step of the flow.};
    \node (labelB) at (6.2, -1.3) {(b) Multi-scale architecture.};

    \end{tikzpicture}
    
    \vspace{0.3cm}
    \caption{An overview of the architecture used to parametrize the reversible simulator. We used $k = 18$ and $\ell=3$. The network is identical to the one proposed in \cite{NEURIPS2018_d139db6a} with the exception that we remove the split layers. For comparison the reader can refer to \cite[Figure 2]{NEURIPS2018_d139db6a}.}
    \label{fig:flow_architecture}
\end{figure}

For training we use a weighted sum $\sum_{i=1}^4 \lambda_i \mathcal{J}_i$ of the losses $\mathcal{J}_1,\dots, \mathcal{J}_4$ with $\lambda_1 = 5$, $\lambda_2 = 10$ and $\lambda_3 = \lambda_4 = 1$ (chosen by trial-and-error). For all losses $\mathcal{J}_i$ we use the maximum mean discrepancy loss with inverse multiquadratics kernel of the form
$$k(\mathbf{p}, \mathbf{q}) = \sum_{i=1}^5 \frac{a_i}{a_i + \lVert \mathbf{p} - \mathbf{q}\rVert_2^2},$$
with $a_1 = 0.1$, $a_2 = 0.5$, $a_3 = 1.0$, $a_4 = 2.0$ and $a_5 = 5.0$. Furthermore, we add a small regularization term on the log determinant of both the forward and reverse pass with a penalizing weight of $10^{-4}$, as otherwise some of the model weights exploded in our experiments. For the optimization we use the ADAM optimizer with a fixed learning rate of $10^{-3}$ for $100$ epochs  and a batch size of $200$.

\subsubsection{Validation}
\begin{figure}[h!]
    \centering
    \includegraphics[width=0.6\textwidth]{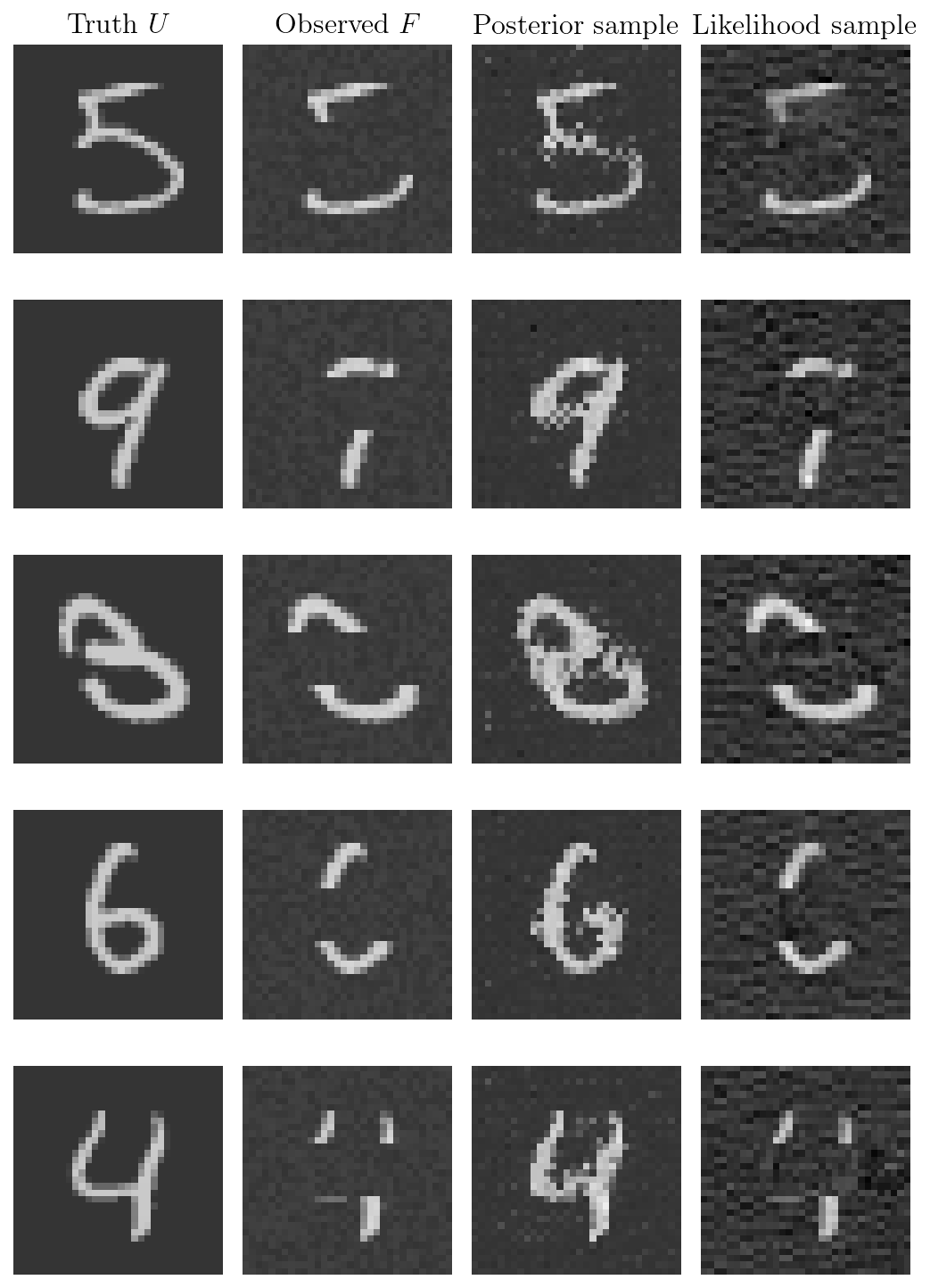}
    \caption{In the first column, the true handwritten digits are shown. In the second, the obscured images. The third column shows a posterior sample, conditioned on the observation in the second column, from the learned reversible simulator. The final column shows a likelihood sample from the learned reversible simulator.}
    \label[figure]{fig: RS inpainting}
\end{figure}

In \cref{fig: RS inpainting} several examples are shown from the MNIST testing dataset. A few things stand out. First, the likelihood samples are quite accurate but the noise appears to be correlated and still carry some imprint of the original digit. Second, the inpainting overall is quite accurate, with the notable exception of the digit '3'. On the other hand, the digit '9' is recovered well, even though it could just as well be consistent with a '7'. 

\begin{figure}[!ht]
    \centering
    \includegraphics[width=0.5\textwidth]{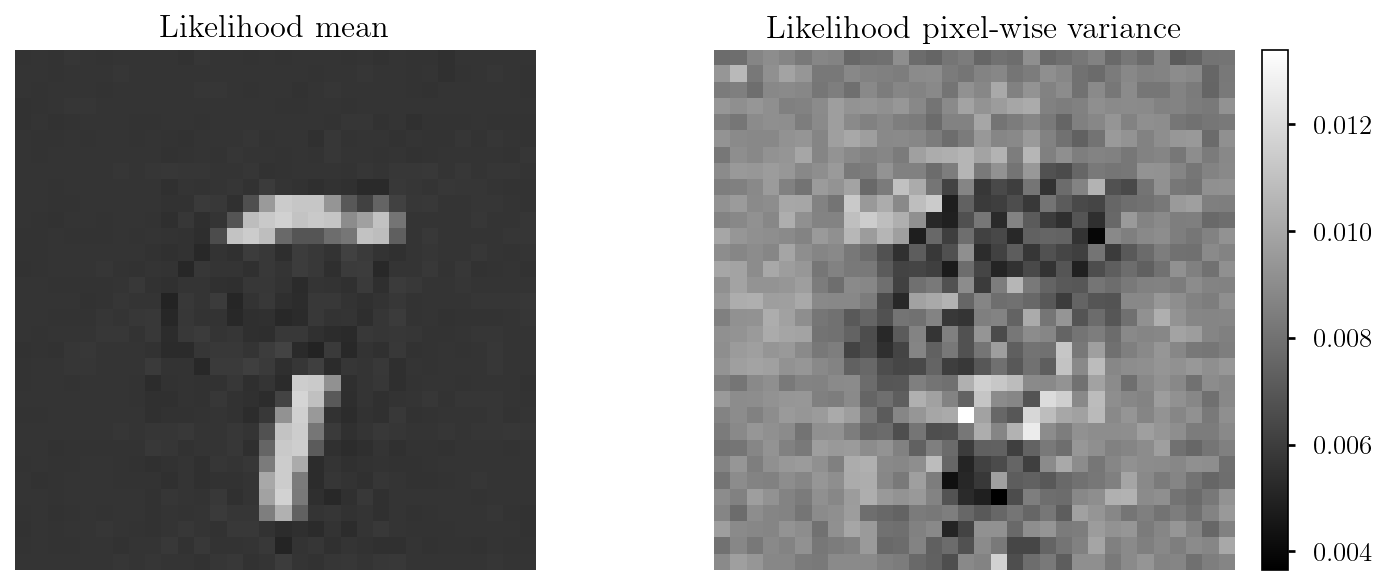}
    \caption{Using the learned reversible simulator, the mean and pixel-wise variance of the likelihood $\mu_{\F\mid \U=\u_9}$ is shown.}
    \label[figure]{fig: RS likelihood mean var}
\end{figure}

\begin{figure}[!ht]
    \centering
    \includegraphics[width=0.5\textwidth]{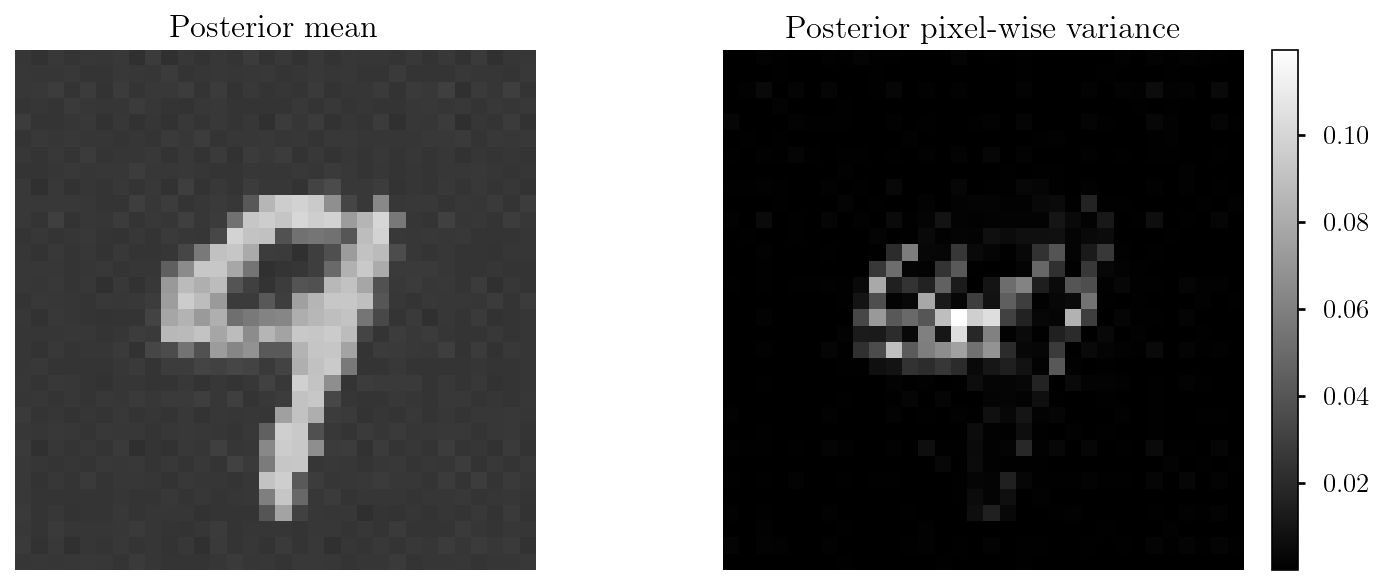}
    \caption{Using the learned reversible simulator, the mean and pixel-wise variance of the posterior $\mu_{\U\mid \F=\f_9}$ is shown.}
    \label[figure]{fig: RS posterior mean var}
\end{figure}

Let us take a closer look at the likelihood and posterior of the digit '9'. In \cref{fig: RS likelihood mean var} we see the mean and pixel-wise standard deviation of generated likelihood samples for the digit '9' shown in \cref{fig: RS inpainting} (first column, second row). This clear shows the remaining imprint of the digit and the non-uniform variance of the noise. In \cref{fig: RS posterior mean var} we see the mean and pixel-wise standard deviation of generated posterior samples for the digit '9' shown in \cref{fig: RS inpainting} (second column, second row). The posterior mean recovers the digit well, and the pixel-wise variance localises the uncertainty to the inpainting region as expected.

% \begin{figure}[!ht]
%     \centering
%     \includegraphics[width=0.8\textwidth]{figures/inpainting/likelihood.png}
%     \caption{Samples from the learned likelihood $\mu_{\F\mid \U=\u_9}$ through the RS are shown, where $\u_9$ is shown in the top left.}
%     \label[figure]{fig: RS likelihood}
% \end{figure}

% In \cref{fig: RS posterior} samples from the approximate posterior $\mu_{\U\mid \F=\f_9}$ are shown. Note that all samples clearly represent the digit $9$, whereas the explicit reversible simulator from our Gaussian prior before was not so sure. Interestingly, in \cref{fig: RS posterior mean var} we do see a high pixel-wise variance in the area where the $9$ in distinct from a potential $7$. So the uncertainty between the $7$ or $9$ solution is not completely gone, but on a sample to sample basis it is not visible, in the sense that on each sample we always get a $9$, just with some pixels missing some of the time.

% \begin{figure}[!ht]
%     \centering
%     \includegraphics[width=0.8\textwidth]{figures/inpainting/posterior.png}
%     \caption{Samples from the learned posterior $\mu_{\U\mid \F=\f_9}$ through the RS are shown, where $\f_9$ is shown in the top left.}
%     \label[figure]{fig: RS posterior}
% \end{figure}

\begin{figure}[!ht]
    \centering
    \includegraphics[width=0.6\textwidth]{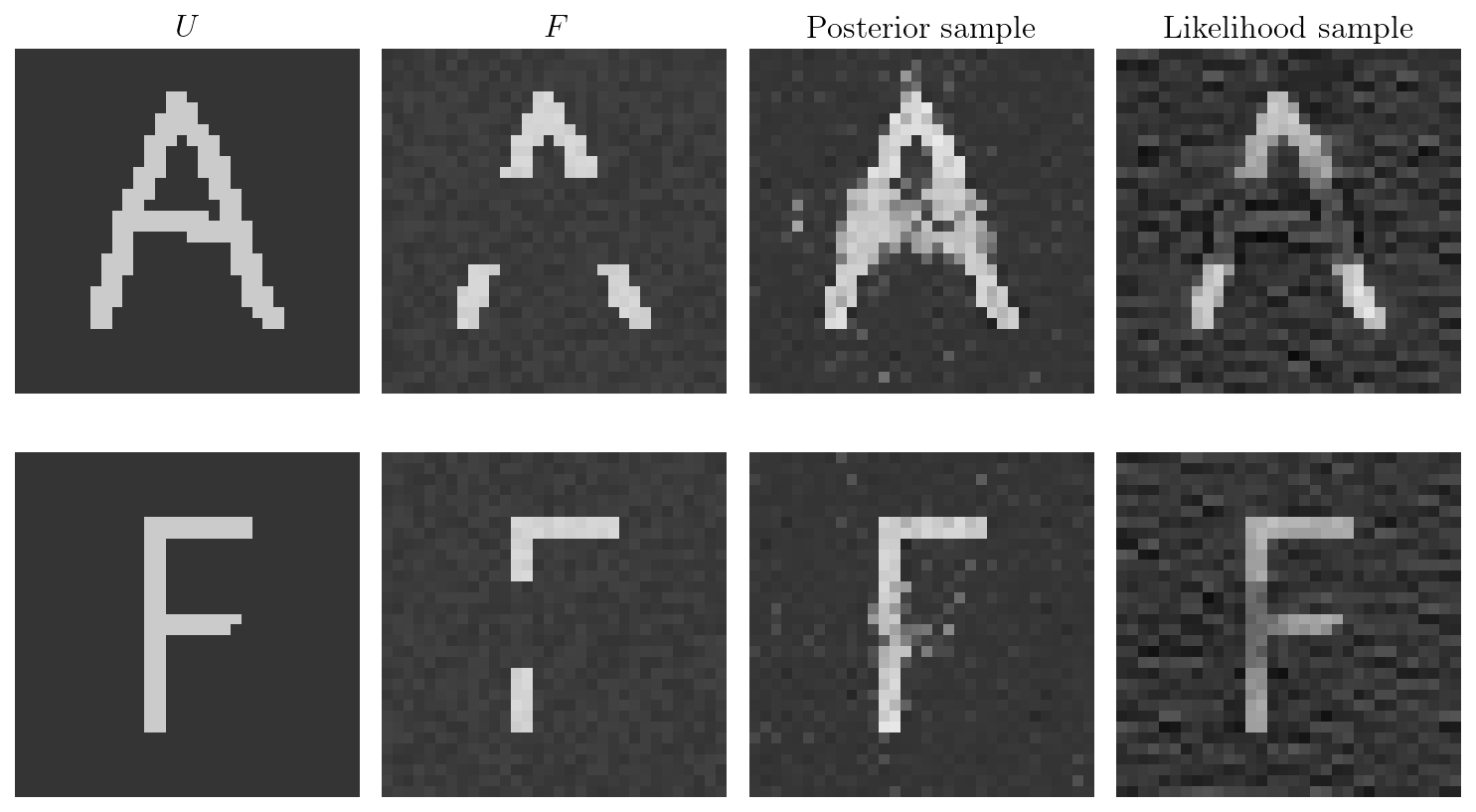}
    \caption{Instead of handwritten digits, the handwritten letters A and F are used.}
    \label[figure]{fig: RS letters}
\end{figure}

Finally, we test the generalization properties of learned reversible simulator to images outside of its training distribution. In particular, in \cref{fig: RS letters} we test how a reversible simulator learned on MNIST performed when it is tested on the letters `A' and `F' under the same inpainting measurement model. The posterior sample recovers the measurement satisfactorily,  keeping in mind that the model has no knowledge of letters. 
Regarding the the likelihood, if the learned reversible simulator had perfectly learned the forward operator, the samples from the likelihood should be perfect, even for images completely outside of its domain. In \cref{fig: RS letters} we see that this is not the case. The pixel values of the image seem to influence the way the image is obstructed, since both letters are being only partly erased on the center.

%%%%%%%%%%%%%%%%%%%%%%%%%%%%%%%%%%%%%%%%%%%%%%%%%%%%%%%%%%%%%%%%%%%%%%%%%%%%%%%%%%%%%%%%%%%%%%%%%%%%%%%%%%%
%%%%%%%%%%%%%%%%%%%%%%%%%%%%%%%%%%%%%%%%%%%%%%%%%%%%%%%%%%%%%%%%%%%%%%%%%%%%%%%%%%%%%%%%%%%%%%%%%%%%%%%%%%%
\section{Conclusion and discussion}\label{conclusion}
%%%%%%%%%%%%%%%%%%%%%%%%%%%%%%%%%%%%%%%%%%%%%%%%%%%%%%%%%%%%%%%%%%%%%%%%%%%%%%%%%%%%%%%%%%%%%%%%%%%%%%%%%%%
%%%%%%%%%%%%%%%%%%%%%%%%%%%%%%%%%%%%%%%%%%%%%%%%%%%%%%%%%%%%%%%%%%%%%%%%%%%%%%%%%%%%%%%%%%%%%%%%%%%%%%%%%%%
We present a generic framework for probabilistic simulation and inference using invertible mappings. The proposed construction has the appealing feature that simulation and inference can be performed with a single generative model, run in either forward or reverse mode. The reversible simulator can be constructed explicitly from lower and upper triangular Knoth-Rosenblath transport maps that push forward a reference distribution to the joint target distribution. The resulting reversible simulator is shown to produce samples from the corresponding conditional distributions (the likelihood and the posterior). Furthermore, we show that the reversible simulator is not unique and that infinitely many can be generated by composition with measure-preserving maps. A variational formulation is proposed, which allows us to learn a reversible simulator from samples. Stylised numerical examples show the workings of the reversible simulator, and how it can be realized using an invertible neural network architecture.

Possible advantages of the proposed construction include the improved conditioning of the mapping that needs to be learned, and the fact that only a single mapping needs to be learned. The improved conditioning is illustrated with a few examples, but needs to be explored further. The current uniqueness result can be probably be strengthened, leading to a characterization of all reversible simulators for given target and reference measures. Such a characterization will open up further avenues for the study of additional optimality criteria (in the spirit of optimal transport) and potentially allow us to enforce this during training in order to get the most favorable (in terms of conditioning) reversible simulator.

While the current work focuses on the construction of the mapping using a transport map, we see opportunities to extent this to other deterministic generative models (such as neural ODEs \cite{NEURIPS2018_69386f6b} and flow matching \cite{lipman2022flow, martin2025pnp}) and also to non-deterministic ones such as stochastic interpolants \cite{albergo2023stochastic} and diffusion models \cite{daras2024survey}. We also aim to explore the use of this framework for experimental design by introducing an additional design parameter, $\mathbf{s}$, over which the model will be amortized during training. The resulting mapping $S(\mathbf{s}) : \mathbb{R}^{m+n}\to\mathbb{R}^{m+n}$ is invertible for every $\mathbf{s}$ and can be used for simulation and inference, and hence for experimental design.
%%%%%%%%%%%%%%%%%%%%%%%%%%%%%%%%%%%%%%%%%%%%%%%%%%%%%%%%%%%%%%%%%%%%%%%%%%%%%%%%%%%%%%%%%%%%%%%%%%%%%%%%%%%
%%%%%%%%%%%%%%%%%%%%%%%%%%%%%%%%%%%%%%%%%%%%%%%%%%%%%%%%%%%%%%%%%%%%%%%%%%%%%%%%%%%%%%%%%%%%%%%%%%%%%%%%%%%
\bibliographystyle{plain}
\bibliography{mybib}
%%%%%%%%%%%%%%%%%%%%%%%%%%%%%%%%%%%%%%%%%%%%%%%%%%%%%%%%%%%%%%%%%%%%%%%%%%%%%%%%%%%%%%%%%%%%%%%%%%%%%%%%%%%
%%%%%%%%%%%%%%%%%%%%%%%%%%%%%%%%%%%%%%%%%%%%%%%%%%%%%%%%%%%%%%%%%%%%%%%%%%%%%%%%%%%%%%%%%%%%%%%%%%%%%%%%%%%

\end{document}